\pdfoutput=1
\documentclass[11pt]{article}

\usepackage{acl}
\usepackage{times}
\usepackage{latexsym}
\usepackage{makecell}

\usepackage[T1]{fontenc}
\usepackage[utf8]{inputenc}
\usepackage{enumitem}
\usepackage{microtype}
\usepackage{amssymb}
\usepackage{amsmath}
\usepackage{booktabs}
\usepackage{multirow}
\usepackage{graphicx}
\usepackage{mathtools, subcaption}
\usepackage{wasysym}
\usepackage{colortbl}

\usepackage{float} 
\usepackage{inconsolata}
\usepackage{xcolor}
\title{Are LLMs Effective Negotiators? Systematic Evaluation of the Multifaceted Capabilities of LLMs in Negotiation Dialogues}

\author{Deuksin Kwon$^1$\hspace{0.3cm}Emily Weiss$^1$\hspace{0.3cm}Tara Kulshrestha$^2$\\\textbf{Kushal Chawla}$^{3}$\hspace{0.3cm}\textbf{Gale M. Lucas}$^1$\hspace{0.3cm}\textbf{Jonathan Gratch}$^1$\\
$^{1}$University of Southern California\hspace{0.3cm}$^2$University of California, Berkeley\hspace{0.3cm}
\\$^3$Capital One\\
\texttt{\{deuksink,emilymwe\}@usc.edu}
\\\texttt{tarakul@berkeley.edu}\hspace{0.3cm}\texttt{kushal.chawla@capitalone.com}
\\\texttt{\{lucas,gratch\}@ict.usc.edu}}

\begin{document}
%\nolinenumbers %Side line numbers
\maketitle

\begin{abstract}
% A successful negotiation demands a deep comprehension of the conversation context, Theory-of-Mind (ToM) skills to infer the partner's motives, as well as strategic reasoning and effective communication, making it challenging for automated systems. Given the remarkable performance of LLMs across a variety of NLP tasks, in this work, we aim to understand how LLMs can advance different aspects of negotiation research, ranging from designing dialogue systems to providing pedagogical feedback and scaling up data collection practices. To this end, we devise a methodology to systemically analyze the multifaceted capabilities of LLMs across diverse dialogue scenarios covering all the time stages of a typical negotiation interaction. Our analysis adds to the increasing evidence for the superiority of GPT-4 across various tasks while also providing insights into specific tasks that remain difficult for LLMs. For instance, the models correlate poorly with human players when making subjective assessments about the negotiation dialogues and often struggle to generate responses that are contextually appropriate as well as strategically advantageous.

A successful negotiation requires a range of capabilities, including comprehension of the conversation context, Theory-of-Mind (ToM) skills to infer the partner’s motives, strategic reasoning, and effective communication, making it challenging for automated systems. Despite the remarkable performance of LLMs in various NLP tasks, there is no systematic evaluation of their capabilities in negotiation. Such an evaluation is critical for advancing AI negotiation agents and negotiation research, ranging from designing dialogue systems to providing pedagogical feedback and scaling up data collection practices. This work aims to systematically analyze the multifaceted capabilities of LLMs across diverse dialogue scenarios throughout the stages of a typical negotiation interaction. Our analysis highlights GPT-4’s superior performance in many tasks while identifying specific challenges, such as making subjective assessments and generating contextually appropriate, strategically advantageous responses. The code is available at \href{https://github.com/DSincerity/SysEval-NegoLLMs}{https://github.com/DSincerity/SysEval-NegoLLMs}

\end{abstract}

\section{Introduction}

%%%%%%%%%%%%%%%%%%%%%%%%%%%%%%%%%%%%%%%%%%%%%%%%%%%
% Intro Figure\begin{figure}[
\begin{figure}[th]
    \centering
    \includegraphics[width=1\linewidth]{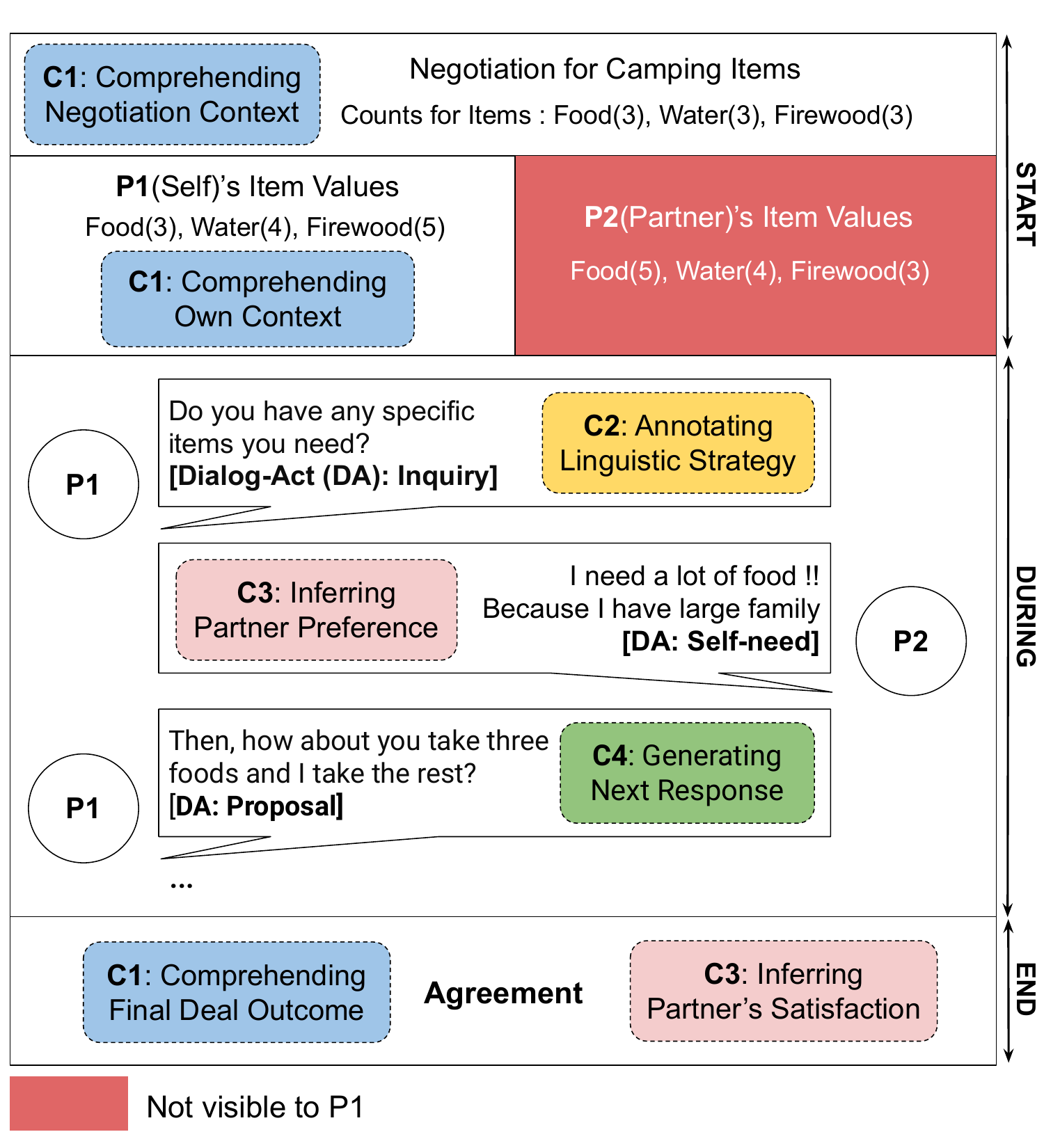}
    \caption{Overview of the key capabilities (C1-C4) required for a successful negotiation. We design tasks aligned with these abilities to assess how LLMs can advance different aspects of negotiation research. The negotiation scenario is based on \citet{chawla-etal-2021-casino}.}
    \label{fig:intro}
\end{figure}

%%%%%%%%%%%%%%%%%%%%%%%%%%%%%%%%%%%%%%%%%%%%%%%%%%

Large Language Models (LLMs), known for their impressive understanding and reasoning capabilities, are increasingly being utilized as AI negotiation agents~\cite{fu2023improving}. However, prior research on LLM-based negotiation agents lacks a systematic analysis and evaluation of the diverse abilities required for effective negotiation~\cite{fu2023improving,gandhi2023strategic,abdelnabi2023llmdeliberation}. To develop more advanced AI negotiating agents, it is crucial to assess the various capabilities necessary for negotiation. Systematic evaluation of LLMs through diverse negotiation tasks can provide interpretable insights by breaking down the complex negotiation process into single tasks. This approach significantly enhances the utility and value of LLMs for various purposes in negotiation dialogues (e.g., designing end-to-end~\cite{lewis2017deal} and modular negotiation dialogue systems~\cite{he2018decoupling}, for providing pedagogical feedback through ToM~\cite{johnson2019intelligent}, and for scaling up data collection practices by automating the collection of utterance-level annotations~\cite{yamaguchi-etal-2021-dialogue}).

For a sophisticated evaluation of the multifaceted capabilities of LLMs in negotiation, it is essential to consider the core competencies required by agents in a realistic 
 yet challenging bilateral multi-issue negotiation scenario~\cite{fershtman1990importance}, as illustrated in Figure\ref{fig:intro}. Players begin with predefined value preferences for certain issues (e.g., food, water, firewood) and engage in dialogue to divide all available items, aiming to maximize their total value. To succeed, a negotiation agent must understand the negotiation context and outcome (\textbf{C1}), grasp the dialogue’s structure (e.g., intent) and semantics (e.g., linguistic strategies) (\textbf{C2}), infer the partner’s preferences through conversation (\textbf{C3}), and generate coherent, strategically apt responses given negotiation context and conversation (\textbf{C4}). These challenges highlight negotiation as a rigorous AI research testbed.

To this end, we propose a comprehensive framework that systematically analyzes state-of-the-art LLMs in negotiations across 35 tasks, considering the varied skills required by competent negotiation agents. Specifically, successful negotiation requires the Theory of Mind (ToM)~\cite{frith2005theory}—the ability to understand the opponent’s perspective—and the capacity to infer subjective aspects like affective states, beyond objective information. Our framework encompasses a range of partner modeling and subjective tasks, defined along three crucial axes: 1) \textit{Objectivity} (Objective vs. Subjective tasks), 2) \textit{Time Stage} of the negotiation (Start, During, and End), and 3) \textit{Task Type} (Comprehension, Annotation, Partner Modeling, and Generation), aligned with the core competencies necessary for successful negotiations (Figure \ref{fig:intro}). Our contributions are as follows:

%%%%%%%%%%%%%%%%%%%%%%%%%%%%%%%%%%%%%%%%%%%%%%%%%%%
% Main Framework Figure
\begin{figure*}[ht]
    \centering
    \includegraphics[width=0.85\linewidth]{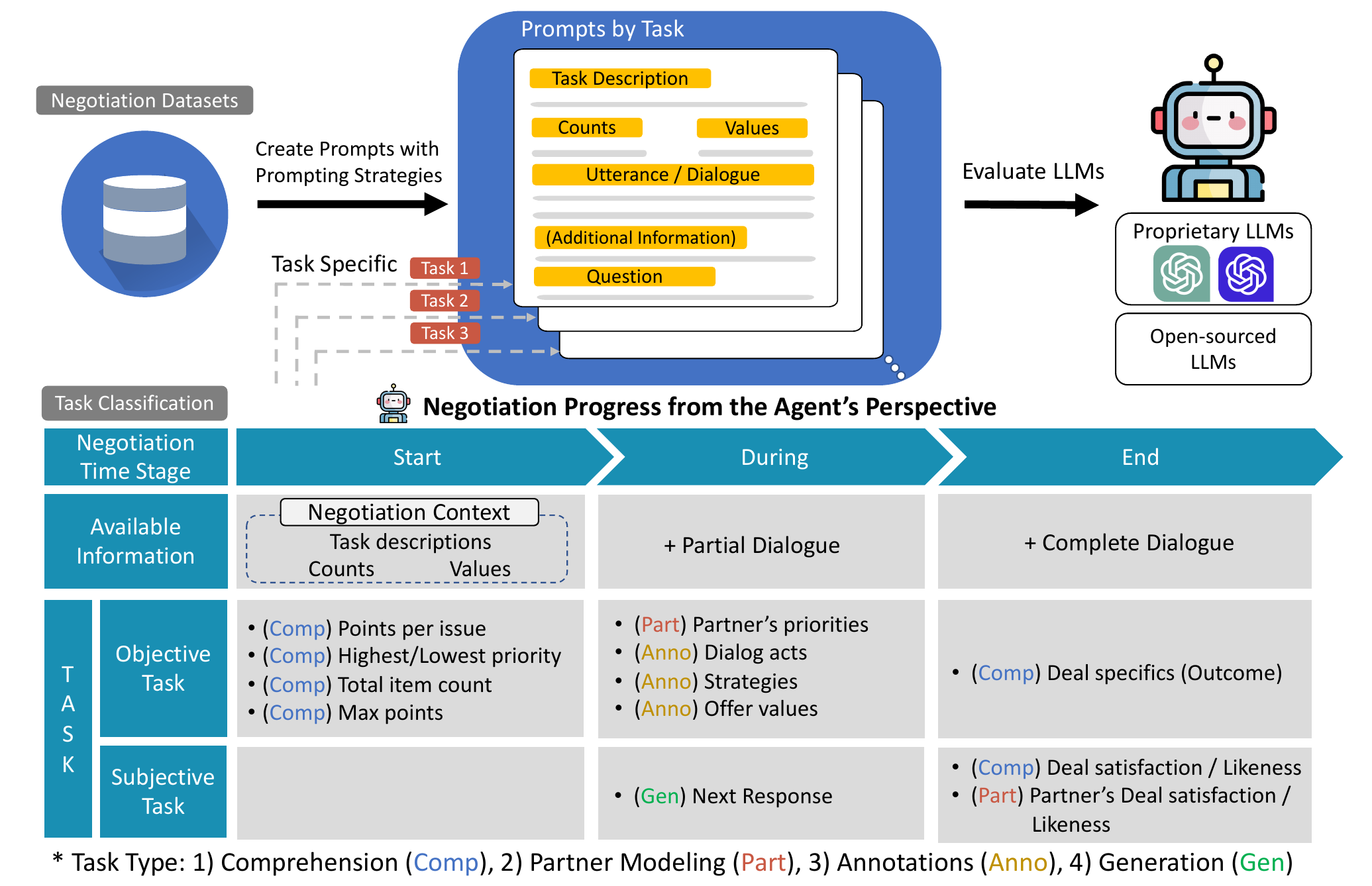}
    \caption{Our methodology for systematically evaluating LLMs in negotiation dialogues. Part A (top) describes the pipeline for creating task-specific prompts from a negotiation dataset and evaluating various LLMs with them. Part B (bottom) depicts the tasks categorized by \textit{Objectivity}, \textit{Time Stage}, and \textit{Task Type} (Section \ref{sec:task-design}).}
    \label{fig:framework}
\end{figure*}
%%%%%%%%%%%%%%%%%%%%%%%%%%%%%%%%%%%%%%%%%%%%%%%%%%
% [NOTE THE NOVELTY AND HOW IT CAN GENERALIZE]  -> do this later -not fitting here.

\begin{enumerate}
\item We design a novel methodology to evaluate the multifaceted capabilities of LLMs in negotiations. While our framework is specifically designed with the goal of supporting negotiation research, the resulting methodology also captures nuances common to other dialogue tasks (Section \ref{sec:methodology}).
\item We evaluate zero-shot out-of-the-box capabilities of LLMs on $35$ tasks based on $4$ dialogue datasets capturing diverse negotiation scenarios. Our overall trends show the superiority of GPT-4, finding that it often outperforms even the Flan-T5 baseline which is specifically fine-tuned for the corresponding task (Section \ref{sec:results}). Additionally, all models show poor correlations with human judgments on subjective assessments about the negotiation.
\item Through an evaluation with expert human annotators, we find that GPT-4 performs comparably to humans in response generation. We perform an error analysis, discussing the Coherence and strategic failures that still persist in the compared LLMs (Section \ref{sec:generation-results}).
\item We uncover intriguing results, finding the effectiveness of CoT and few-shot prompting. Also, by evaluating LLMs’ prediction balance, we confirm GPT-4’s superior robustness compared to other models (Section \ref{sec:discussion}).
% \item We uncover some additional intriguing results. We explore the impact of various prompt strategies, discovering the effectiveness of CoT and few-shot prompting. Also, by evaluating LLMs’ prediction balance, we confirm GPT-4’s superior robustness compared to other models (Section \ref{sec:discussion}).
%\item We explore the impact of various prompting strategies and provide recommendations for future work (Sections \ref{sec:discussion} and \ref{sec:conclusion}).
\end{enumerate}

\section{Methodology}
\label{sec:methodology}

We evaluate the multifaceted capabilities of LLMs in negotiations with a three-step framework (Figure \ref{fig:framework}): 1) Designing tasks from human-human dialogue datasets, 2) Constructing task-specific prompts, and 3) Evaluating and analyzing various proprietary and open-source LLMs.

\subsection{Datasets}
\label{subsec:datasets}

Three key requirements guided our dataset selections: 1) Well-structured design with concrete player preferences and outcomes, 2) Diversity in task scenarios, and 3) Availability of metadata for testing holistic LLM abilities. Based on these criteria, we selected four datasets: \textit{CRA} \cite{devault2015toward}, \textit{DND} \cite{lewis2017deal}, \textit{CA} \cite{chawla-etal-2021-casino}, and \textit{JI} \cite{yamaguchi-etal-2021-dialogue}. Table \ref{tab:datasets} provides an overview of these datasets. Each dataset captures a distinct scenario but is based on the Multi-Issue Bargaining Task (MIBT) framework~\cite{fershtman1990importance}, a standard form used for negotiation research, which involves bilateral negotiations over fixed quantities of issues (Table \ref{tab:datasets}). Players have predefined preferences, or \textit{values}, for each issue, which establish their priority order (ex., Scores: 5 = high, 4 = medium, 3 = low priority). In line with standard practice for experimental settings analyzing final outcomes and player behavior, these preferences remain fixed throughout the negotiation.

In MIBT, players engage in dialogue to divide available items to maximize their total points, computed by the inner product of preferences and acquired items in the final deal. MIBT captures various real-world negotiations and has been widely adopted in NLP, beyond NLP~\cite{mell2017grumpy, acharya2013delays, lau2008mining}, and industry as well (e.g., iDecisionGames\footnote{\url{https://idecisiongames.com/promo-home}}).

Some datasets, like \textit{CA}, report subjective metrics such as \textit{satisfaction from the outcome} and \textit{likeness towards their partner}. Others include utterance-level dialogue act (e.g., \textit{CRA}) and strategy (e.g., \textit{CA}) annotations. We use this metadata to design a variety of tasks to evaluate LLMs.

%%%%%%%%%%%%%%%%%%%%%%%%%%%%%%%%%%%%%%%%%%%%%%%%%%
% Dataset Table
\begin{table}[t!]
\centering
\resizebox{\linewidth}{!}{%
\begin{tabular}{@{}cccr@{}}
Dataset & Scenario   & Issues                                                                                     & \multicolumn{1}{c}{\# of Dialogues} \\ \midrule
CRA     & Artifacts trading     & (Painting, Lamp, Album)                                                                    & 119                                                        \\
DND     &  General items  & (Ball, Hat, Book)                                                                          & 6,251                                                      \\
CA      & Campsite Neighbors  & (Food, Water, Firewood)                                                                    & 1,030                                                      \\
JI      & Job Recruiter-Worker & \begin{tabular}[c]{@{}c@{}}(Salary, Day-off, Position, \\ Company, Workplace)\end{tabular} & 2,639                                                      \\
\end{tabular}}
\caption{The datasets used in our analysis. Depending on the dataset and task, we sample and build test instances per task (capped at 200) at either the utterance level or the dialogue level for our evaluation. The training data is used for fine-tuning our baseline model.}
\label{tab:datasets}
\end{table}
%%%%%%%%%%%%%%%%%%%%%%%%%%%%%%%%%%%%%%%%%%%%%%%%%%

\subsection{Task Design}
\label{sec:task-design}

We build 35 tasks based on the aforementioned datasets, guided by the principle that \textit{Every task is designed from the perspective of an agent negotiating for itself}. This is crucial as it governs the information used in the prompts (e.g., explicit partner preferences are not visible).

To categorize these tasks systematically, we use three criteria: 1) \textit{Objectivity}, 2) \textit{Time stage}, and 3) \textit{Task Type} (Figure \ref{fig:framework}). While tailored for specific negotiation scenarios, these criteria also apply to traditional task-oriented and open-domain dialogues, making our evaluation framework broadly applicable. Detailed descriptions of each task are provided in Table \ref{tab:task_desc_of_all_tasks} in the Appendix~\ref{append:Details_of_Negotiation_Tasks}.

\subsubsection{Objectivity}
\label{Subjective_tasks}

Most tasks are \textit{objective}, testing whether models can accurately answer questions about the objective facts of a negotiation and are evaluated using standardized metrics. In contrast, a few tasks are \textit{subjective} where there is no one correct answer. These include predicting outcome satisfaction and partner likability as reported by human players in post-surveys. Subjective measures are crucial for successful task outcomes, especially in repeated interactions between stakeholders, as they affect user perception of agents ~\cite{oliver1994outcome, mell2019likeability, chawla2021towards}. Therefore, we design tasks to assess whether model predictions align with human-reported outcomes by computing correlations. Response generation is also considered as a subjective task, evaluated through automatic and human assessments.

\subsubsection{Time Stage}
\label{time_stage}

A negotiation dialogue typically unfolds in three stages: \textit{Start}, \textit{During}, and \textit{End}. This distinction impacts the information used in prompts, as an agent can only access information available at each stage from its own perspective.

At the \textit{start}, before any dialogue occurs, the agent knows only the basic information from the negotiation context, including the scenario description, item counts, and priority values (Figure \ref{fig:framework}). \textit{During} the dialogue, the agent has access to the partial historical conversation. At the \textit{end}, tasks use the entire dialogue within their prompts.

\subsubsection{Task Type}
\label{task_cate}
Based on prior research, we categorize all subtasks into four types  (i.e., Comprehension~\cite{twitchell2013negotiation, nouri2013prediction}, Annotation~\cite{heddaya2023language}, Generation~\cite{lewis2017deal}, and Partner modeling~\cite{zhang2020learning, chawla2022opponent}), aligning with traditional higher-level modeling tasks.

\noindent \textbf{Comprehension}: These tasks assess the model’s ability to understand the negotiation context and outcome, which is fundamental for any sensible negotiation agent~\cite{cao2015automated}. In the \textit{Start} stage, tasks include identifying the total number of items, issue priorities, and maximum points the agent can gain. In the \textit{End} stage, they evaluate the model’s understanding of the final outcomes, including the final deal details~\cite{twitchell2013negotiation} and subjective measures of satisfaction and partner likability~\cite{curhan2010objective}. Proficiency in these tasks aids in automatically evaluating human-human or human-agent negotiations.

% \noindent \textbf{Comprehension}: In the \textit{Start} stage, these tasks assess the model's ability to understand the negotiation context, which is fundamental for any sensible negotiation agent~\cite{cao2015automated}. This includes asking the total number of available items, the priorities for the issues, and the maximum points the agent can earn from the negotiation. In the \textit{End} stage, Comprehension tasks evaluate whether the models can answer questions about the final outcomes, including details of the final deal~\cite{twitchell2013negotiation} and the subjective measures of satisfaction and partner likeness~\cite{curhan2010objective}. Proficiency in these tasks can aid in automatically evaluating human-human or human-agent negotiations.

\noindent \textbf{Annotation}: These tasks involve annotating utterances with their semantics and functions, such as dialogue acts (e.g., disagree, propose) and linguistic strategies (e.g., elicit-preference, self-need), as well as parsing incoming offers~\cite{chawla-etal-2021-casino}. Annotation is crucial for the Natural Language Understanding (NLU) module in a modular dialogue system, especially \textit{during} the negotiation to understand the partner’s utterances and decide the agent’s next actions.

\noindent \textbf{Partner Modeling}: This is vital \textit{during} the negotiation for understanding the other party’s strategy and priorities, and adapting accordingly for favorable outcomes. These tasks are closely related to evaluating the ToM ability of LLMs in inferring the mental states of the partner during negotiations. We focus on predicting the partner’s priorities and subjective assessments, such as satisfaction with an outcome and likability towards the other party.

% \noindent \textbf{Partner Modeling}: Partner Modeling is particularly vital \textit{during} the negotiation, where understanding the other party's strategy and priorities, and adapting one's strategy accordingly can be key for efficient and favorable outcomes. These tasks are closely related to evaluating the ToM ability of LLMs in inferring the mental states of the partner during negotiations. In our work, we focus on predicting the partner's priorities and subjective assessments (e.g., satisfaction towards an outcome, likeness towards the other party). 

\noindent \textbf{Generation}: This task involves generating responses based on context and dialogue history, including crafting offers and counteroffers, responding to incoming offers, and other communicative elements that advance the negotiation. This is an essential requirement for a conversational agent and tests whether the models can reason through the current state and respond in a way that is both \textit{contextually appropriate} and \textit{strategically advantageous}.

\subsection{Building Task-specific Prompts}

To assess LLMs’ capabilities on our proposed tasks, we create task-specific prompts using a standardized template. As shown in Figure \ref{fig:framework}, the zero-shot prompt template includes five elements: 1) task description, 2) issue counts and values, 3) dialogue or utterance, 4) additional information, and 5) the question.  For each task, each element is filled with relevant information from the dataset instance to complete the task-specific prompt. The inclusion of dialogue and utterances depends on the time stage and task definition, while additional information like dialogue acts and strategy types is included only for annotation tasks.

For our analysis in Section \ref{sec:discussion}, we also explore alternative prompting strategies, such as few-shot prompting and Chain-of-Thought (CoT)~\cite{wei2023chainofthought}. Detailed task descriptions, including questions and applicable datasets, and prompt examples are provided in Appendix  \ref{append:Details_of_Negotiation_Tasks} and \ref{append:prompting}, respectively.

\section{Experiment Design}
% Baseline model 
Our primary goal is to analyze the effectiveness of state-of-the-art LLMs for strategic negotiation interactions. Here, we discuss the compared models and evaluation metrics for the results presented in Section \ref{sec:results} (overall zero-shot results) and Section \ref{sec:discussion} (task-specific analysis to gain further insights).

\subsection{Baselines}
We use the Majority-label voting model and a taskwise fine-tuned Flan-T5 (base)~\cite{chung2022scaling} as baselines. Flan-T5 was chosen for its strong performance across various NLP tasks and its flexibility in handling diverse input and output formats. For fine-tuning, we use the AdamW optimizer~\cite{loshchilov2019decoupled} with an initial learning rate of $5$e-$5$ and a linear scheduler. Models are trained for up to $5$ epochs with a batch size of $8$, selecting the best-performing checkpoint for evaluation. Detailed information about the training process, including data processing steps and computational resources, is provided in Appendix \ref{append:details_bs_training}.

%%%%%%%%%%%%%%%%%%%%%%%%%%%%%%%%%%%%%%%%%%%%
% This is 3 X 2 figures
%%%%%%%%%%%%%%%%%%%%%%%%%%%%%%%%%%%%%%%%%%%%
% Task Types
\begin{figure*}[htbp]
  \centering
  \begin{subfigure}[b]{0.31\textwidth}
    \includegraphics[width=\textwidth, height=3.4cm]{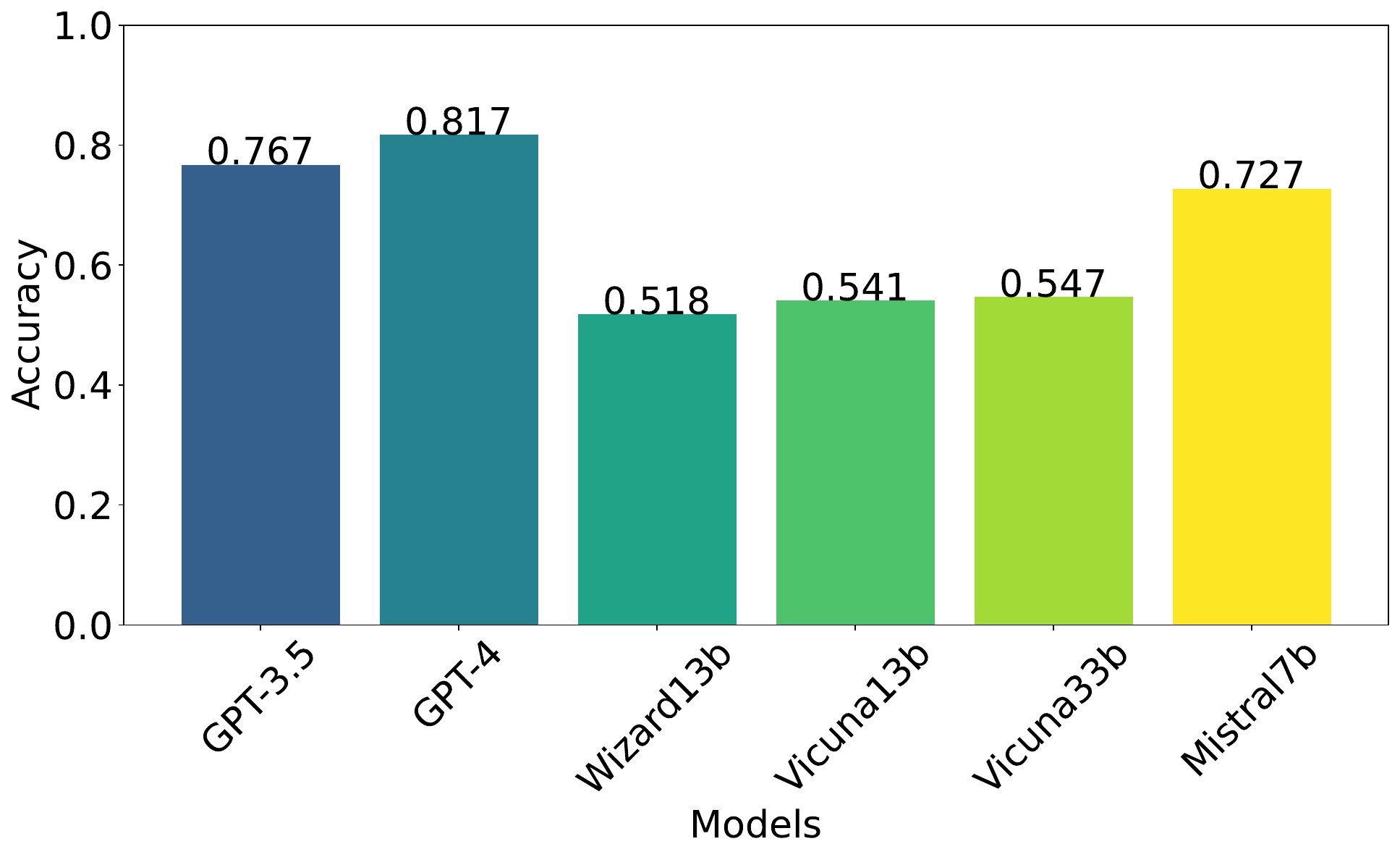}
    \caption{Comprehension (Start) Tasks}
    \label{fig:main_a}
  \end{subfigure}
  \begin{subfigure}[b]{0.31\textwidth}
    \includegraphics[width=\textwidth, height=3.4cm]{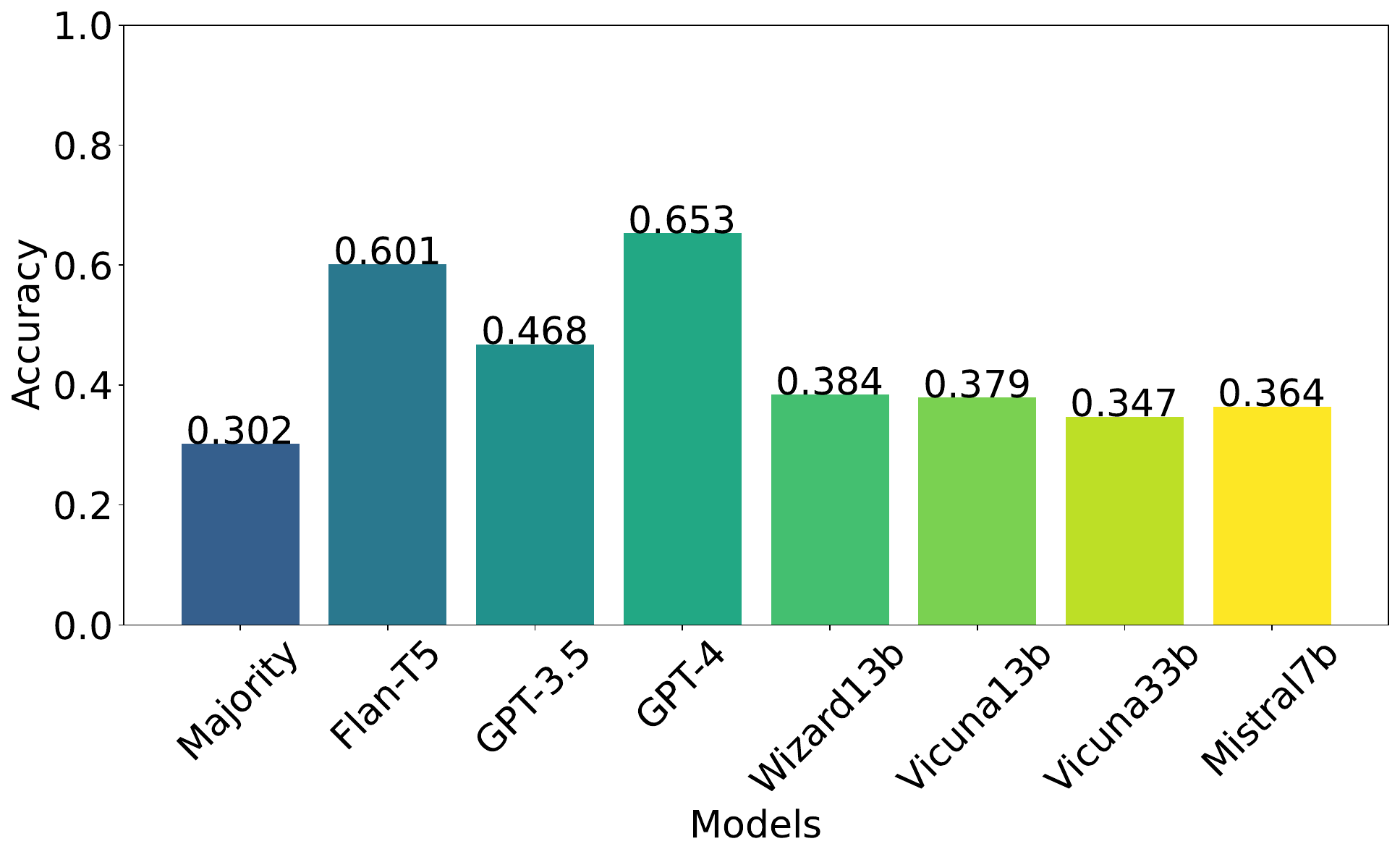}
    \caption{Comprehension (End) Tasks}
    \label{fig:main_b}
  \end{subfigure}
  \begin{subfigure}[b]{0.31\textwidth}
    \includegraphics[width=\textwidth, height=3.4cm]{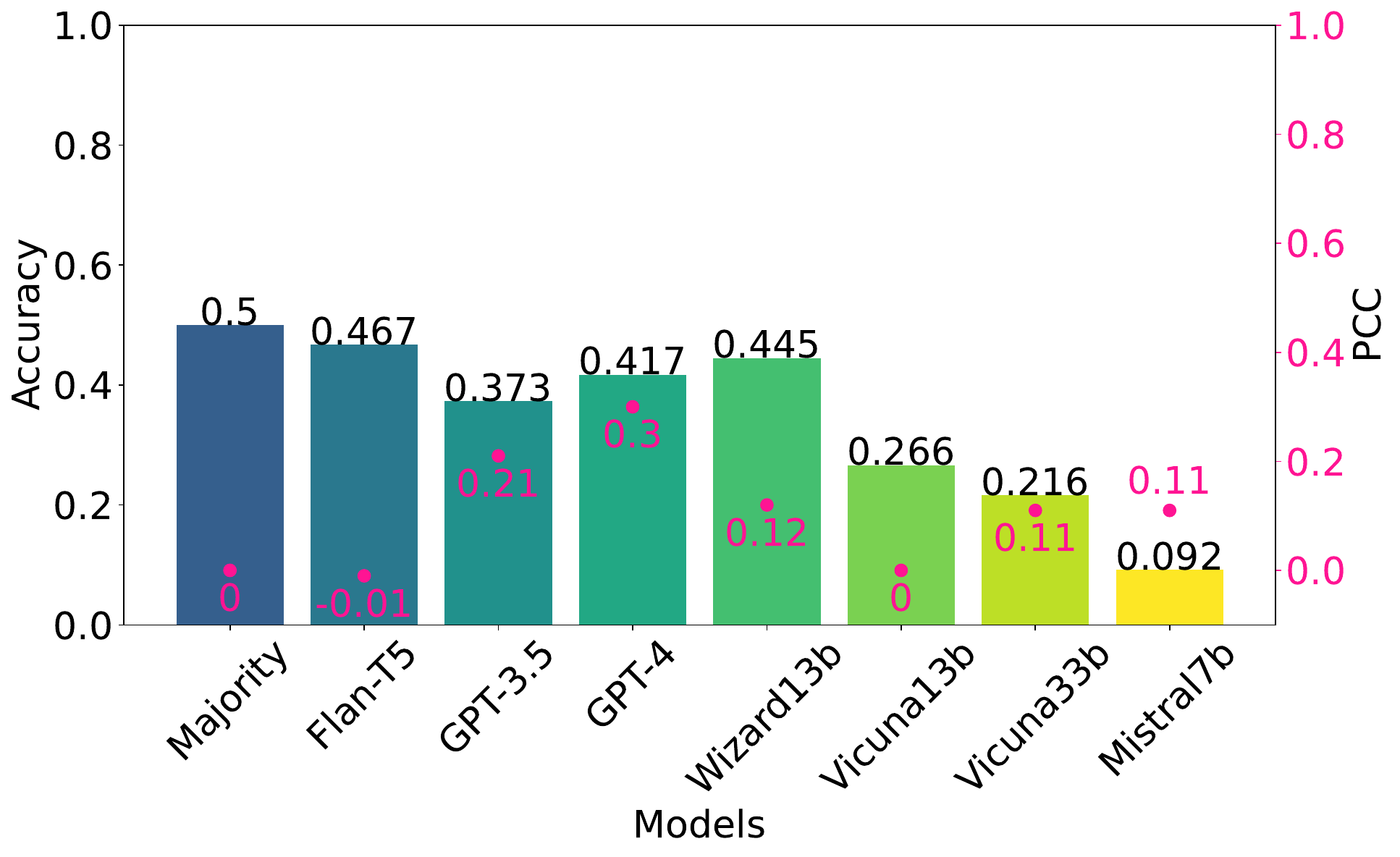}
    \caption{Comprehension (Subjective) Tasks}
    \label{fig:main_c}
  \end{subfigure}
  
  \begin{subfigure}[b]{0.31\textwidth}
    \includegraphics[width=\textwidth, height=3.4cm]{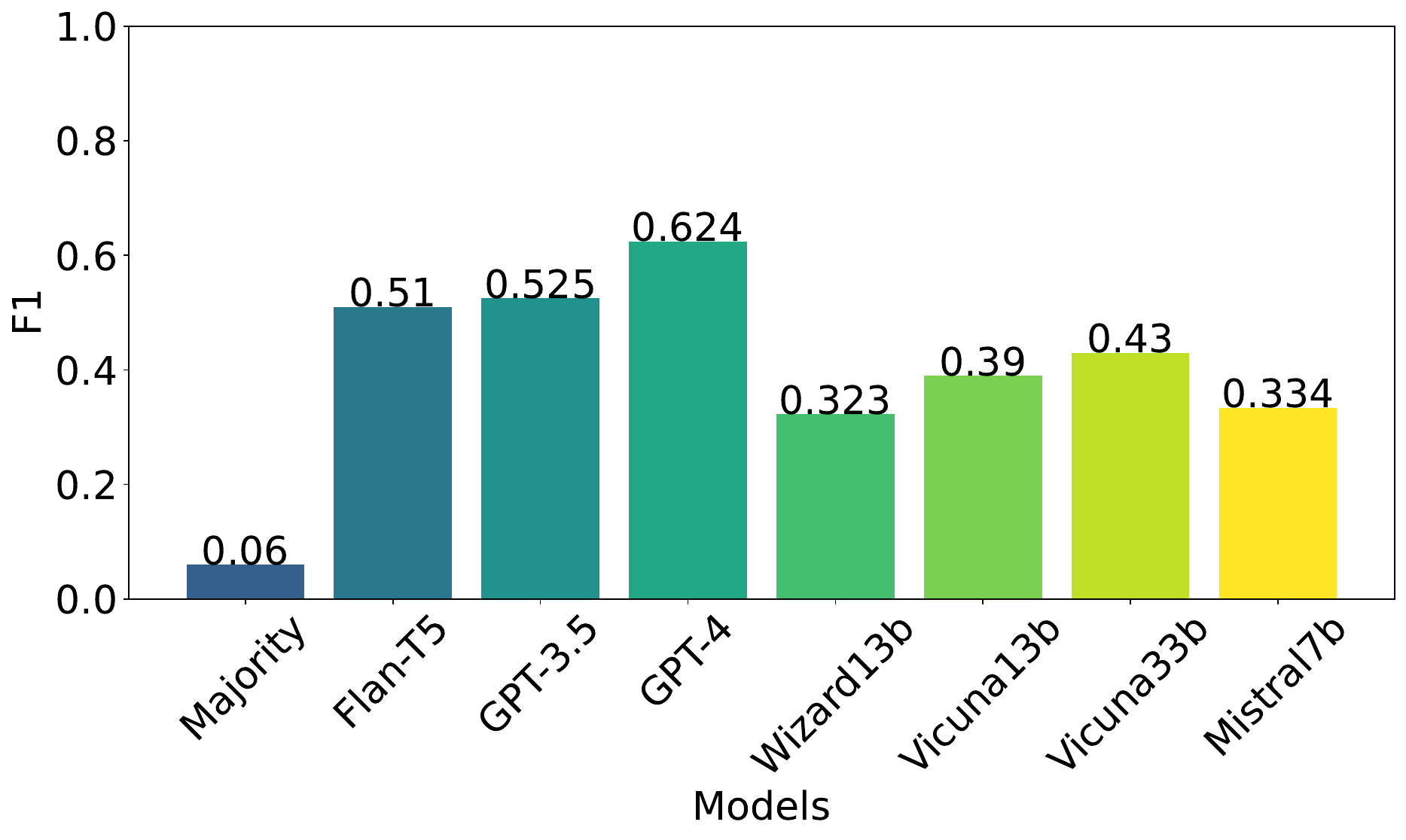}
    \caption{Annotation Tasks}
    \label{fig:main_d}
  \end{subfigure}
  \begin{subfigure}[b]{0.31\textwidth}
    \includegraphics[width=\textwidth, height=3.4cm]{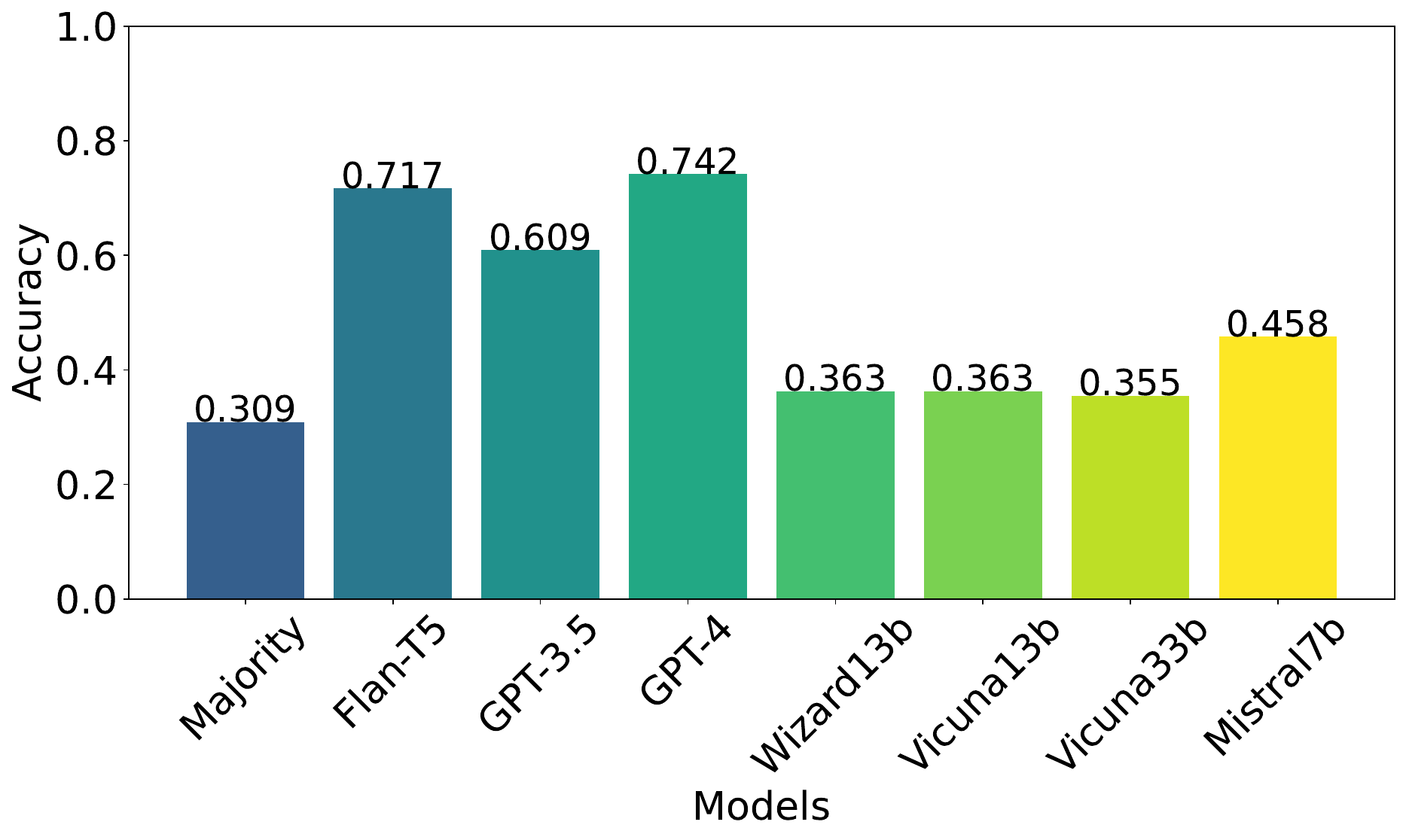}
    \caption{Partner Modeling Tasks}
    \label{fig:main_e}
  \end{subfigure}
  \begin{subfigure}[b]{0.31\textwidth}
    \includegraphics[width=\textwidth, height=3.4cm]{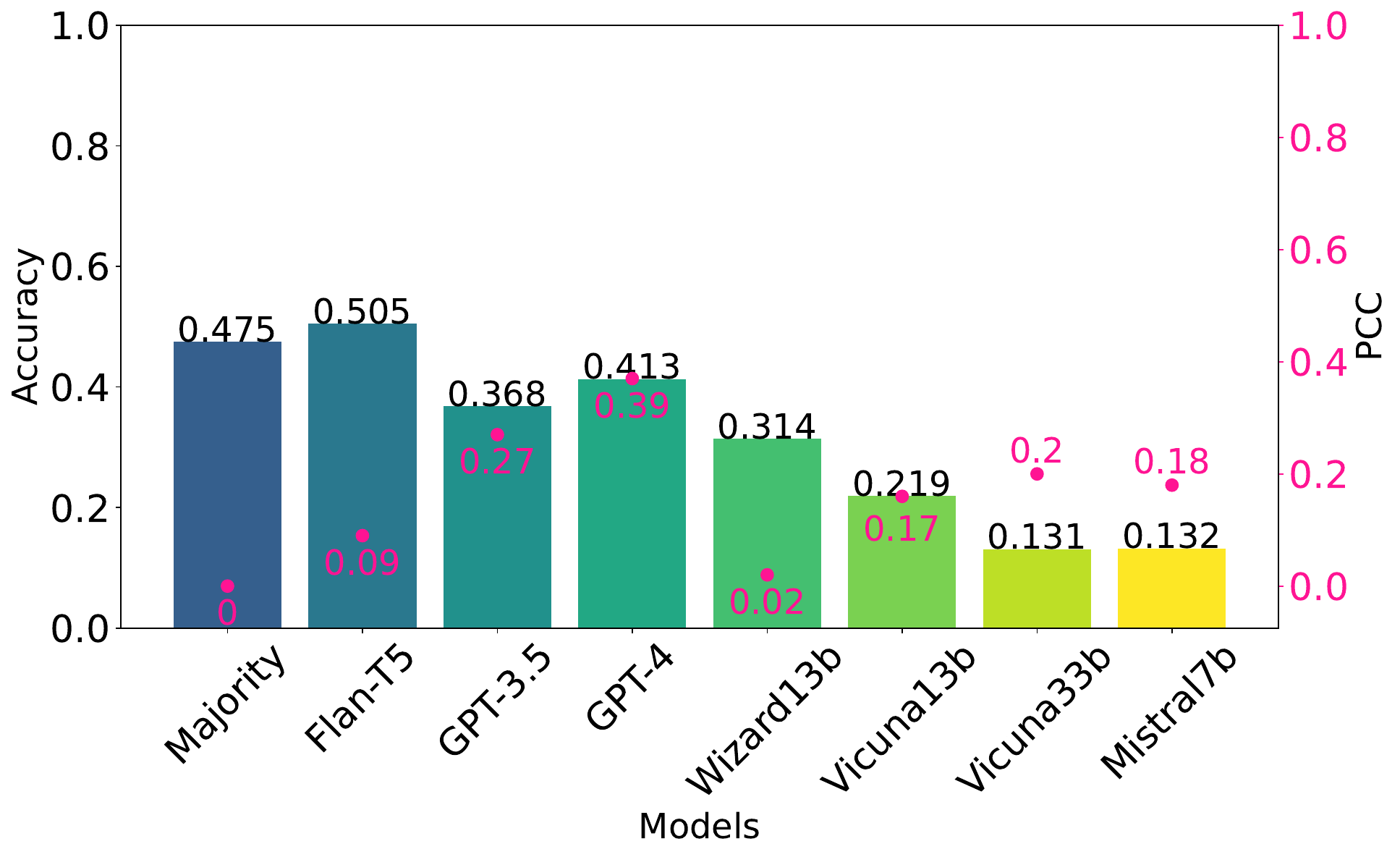}
    \caption{Partner Modeling (Subjective) Tasks}
    \label{fig:main_f}
  \end{subfigure}
  \caption{Overall results for zero-shot evaluation of LLMs. F1: macro F1 over all labels, PCC: Pearson Correlation Coefficient. Each bar shows the average result across all suitable tasks in the category. For example, as per (b), GPT-4 gets $65.3\%$ Accuracy on average for \textit{Comprehension} tasks in \textit{End} time stage. The tasks for these plots have been carefully selected to ensure a fair comparison, with all models passing generation validity checks (i.e., without null values across models), and details of validity check and full results are in Table \ref{tab:full_results_all_models} of Appendix \ref{append:Details_of_Negotiation_Tasks}.}
  \label{fig:result_task_type}
\end{figure*}
%%%%%%%%%%%%%%%%%%%%%%%%%%%%%%%%%%%%%%%%%%%%%%%%%%%%%%%%%%%%%%%%

\subsection{LLMs for evaluation}
%\noindent \textbf{LLMs for evaluation} 
% This study evaluates the multifaceted capabilities of various LLMs as negotiation agents.

We compare a variety of LLMs, top-performing on popular leaderboards at the time of experimentation. Among proprietary models, we choose OpenAI's GPT-3.5-Turbo~\cite{gpt3openai2022} and GPT-4~\cite{gpt4openai2023}. Among open-source LLMs, we experiment with Mistral-7B~\cite{jiang2023mistral} and Wizard-13B~\cite{xu2023wizardlm}, along with Vicuna 13B and 33B variants~\cite{vicuna2023}.

%To obtain the model's output for the task, we utilize the APIs provided by OpenAI for models in the GPT series, while the results for the open-sourced models are inferred from publicly available model files.

\subsection{Evaluation Metrics}

The metrics used depend on the task type. For most tasks, primarily Comprehension and Partner modeling tasks, we rely on Accuracy. For annotation tasks, we use Macro-F1 to account for label imbalance. For subjective tasks measuring satisfaction and likeness scores, we use the Pearson Correlation Coefficient (PCC), along with Accuracy. For response generation, we report automatic evaluation with BLEU, ROUGE, and BERTScore\footnote{https://huggingface.co/google-bert/bert-base-uncased}~\cite{zhang2019bertscore}, and human evaluation by five expert annotators using two subjective metrics on a scale of 1-5: \textit{Coherence} (How appropriate is the response given the dialogue history?) and \textit{Strategy} (How strategic is the response given the agent context and the goal of maximizing performance?). The metrics used for the tasks are detailed in Table \ref{tab:full_results_all_models} in Appendix \ref{append:Details_of_Negotiation_Tasks}.

\section{Results}
\label{sec:results}

Figure \ref{fig:result_task_type} and Table \ref{tab:generation_results} present the key overall trends for zero-shot evaluation on the test set. As detailed in the appendix \ref{append:power_analysis}, our test set is statistically powered and large enough to ensure reliable comparisons across models. We provide the complete task-wise results in Appendix \ref{append:Details_of_Negotiation_Tasks} and discuss the key findings below.

% \begin{table*}[hbt]
% \centering
% \resizebox{0.8\linewidth}{!}{%
% \begin{tabular}{@{}l|llll|llll@{}}
% \multirow{2}{*}{Model} & \multicolumn{4}{c|}{DND} & \multicolumn{4}{c}{CA} \\
% & BLEU$\uparrow$ & Rouge-L$\uparrow$ & Coherence$\uparrow$ & Strategy$\uparrow$ & BLEU$\uparrow$ & Rouge-L$\uparrow$ & Coherence$\uparrow$ & Strategy$\uparrow$ \\ \midrule
% Human & & & $\mathbf{4.5}$ & $\mathbf{4.39}$ & & & $\mathbf{4.14}$ & $\mathbf{3.38}$ \\
% Flan-T5 & $\mathbf{.167}$ & $\mathbf{.453}$ & $4.26^*$ & $4.18$ & $\mathbf{.028}$ & $\mathbf{.165}$ & $3.21^*$ & $2.79^*$ \\ \midrule
% Mistral7b & $.010$ & $.130$ & $3.48^*$ & $2.96^*$ & $.010$ & $.130$ & $2.99^*$ & $2.68^*$\\
% Wizard13b & $.032$ & $.190$ & $3.14^*$ & $3.01^*$ & $.017$ & $.135$ & $3.08^*$ & $2.88^*$\\
% Vicuna13b & $.022$ & $.172$ & $3.48^*$ & $3.34^*$ & $.015$ & $.135$ & $3.36^*$ & $2.92^*$\\
% Vicuna33b & $.038$ & $.216$ & $3.86^*$ & $3.74^*$ & $.016$ & $.147$ & $3.96$ & $3.06^*$\\ \midrule
% GPT-3.5 & $.030$ & $.200$ & $3.8^*$ & $3.50^*$ & $.025$ & $.162$ & $3.60^*$ & $3.01^*$\\
% GPT-4 & $.017$ & $.178$ & $4.47$ & $4.04^*$ & $.011$ & $.149$ & $4.05$ & $3.24$\\

% \end{tabular}}
% \caption{Results on response generation. BLEU and Rouge-L are computed on the full test set ($200$ instances). Coherence and Strategy are based on a human evaluation of a random subset of $50$ examples. $^*$ means significantly worse performance than the Human reference according to a Paired Bootstrap test.}
% \label{tab:generation_results}
% \end{table*}

\begin{table*}[hbt]
\centering
\resizebox{0.85\linewidth}{!}{%
\begin{tabular}{@{}l|lllll|lllll@{}}
\multirow{2}{*}{Model} & \multicolumn{5}{c|}{DND}                                                                                                                                                                           & \multicolumn{5}{c}{CA}                                                                                                                                                                        \\
& \multicolumn{1}{c}{BLEU$\uparrow$} & \multicolumn{1}{c}{Rouge-L$\uparrow$} & \multicolumn{1}{c}{BERTScore$\uparrow$}     & \multicolumn{1}{c}{Coherence$\uparrow$} & \multicolumn{1}{c|}{Strategy$\uparrow$} & \multicolumn{1}{c}{BLEU$\uparrow$} & \multicolumn{1}{c}{Rouge-L$\uparrow$} & \multicolumn{1}{c}{BERTScore$\uparrow$} & \multicolumn{1}{c}{Coherence$\uparrow$} & \multicolumn{1}{c}{Strategy$\uparrow$} \\ \midrule
Human                  & \multicolumn{1}{l}{}               & \multicolumn{1}{l}{}                  & \multicolumn{1}{l}{}              & $\mathbf{4.5}$                          & $\mathbf{4.39}$                         &                                    &                                       &                               & $\mathbf{4.14}$                         & $\mathbf{3.38}$                        \\
Flan-T5                & $\mathbf{.167}$                    & $\mathbf{.453}$                       & $\mathbf{.678}$ & $4.26^*$                                & $4.18$                                  & $\mathbf{.028}$                    & $\mathbf{.165}$                       & $.468$                         & $3.21^*$                                & $2.79^*$                               \\ \midrule
Mistral7b              & $.010$                             & $.130$                                & $.401$                             & $3.48^*$                                & $2.96^*$                                & $.010$                             & $.130$                                & $.401$                         & $2.99^*$                                & $2.68^*$                               \\
Wizard13b              & $.032$                             & $.190$                                & $.451$                             & $3.14^*$                                & $3.01^*$                                & $.017$                             & $.135$                                & $.466$                         & $3.08^*$                                & $2.88^*$                               \\
Vicuna13b              & $.022$                             & $.172$                                & $.486$                             & $3.48^*$                                & $3.34^*$                                & $.015$                             & $.135$                                & $.472$                         & $3.36^*$                                & $2.92^*$                               \\
Vicuna33b              & $.038$                             & $.216$                                & $.547$                             & $3.86^*$                                & $3.74^*$                                & $.016$                             & $.147$                                & $.483$                         & $3.96$                                  & $3.06^*$                               \\ \midrule
GPT-3.5                & $.030$                             & $.200$                                & $.467$                             & $3.8^*$                                 & $3.50^*$                                & $.025$                             & $.162$                                & $\mathbf{.495}$              & $3.60^*$                                & $3.01^*$                               \\
GPT-4                  & $.017$                             & $.178$                                & $.489$                             & $4.47$                                  & $4.04^*$                                & $.011$                             & $.149$                                & $.48$                          & $4.05$                                  & $3.24$                                
\end{tabular}}
\caption{Results on response generation. BLEU, Rouge-L and BERTScore are computed on the full test set ($200$ instances). Coherence and Strategy are based on a human evaluation of a random subset of $50$ examples. $^*$ means significantly worse performance than the Human reference according to a Paired Bootstrap test~\cite{sakai2006evaluating}.}
\label{tab:generation_results}
\end{table*}

\subsection{Comprehension Tasks}

As shown in Figure \ref{fig:main_a}, GPT-4 outperforms all other models on Comprehension tasks in the \textit{Start} stage, with an average accuracy of 81\%. This is followed by GPT-3.5 and Mistral7B, both scoring above 70\%. However, these tasks primarily involve questions about the explicitly provided negotiation context, so a simple rule-based parser could achieve 100\% accuracy, indicating significant room for improvement.

The \textit{End} stage tasks are more challenging, requiring models to comprehend the dialogue, reason to extract the agreed deal, and compute the answer. As shown in Figure \ref{fig:main_b}, most models fail to perform well, only marginally better than the trivial Majority baseline. GPT-4 still demonstrates superior performance, surpassing Flan-T5. In comparison, since all evaluated instances end in an \textit{unambiguous} final deal, an expert human can achieve nearly perfect scores on these tasks.

For subjective tasks, we analyze if the LLMs’ self-assessments of outcome satisfaction and partner likeness align with those reported by human players. Figure \ref{fig:main_c} shows that all models perform poorly in terms of Accuracy and PCC. Although GPT-4 exhibited relatively better performance, it sometimes misclassified deal satisfaction in a completely opposite way to humans (e.g., classifying satisfied deals as dissatisfied, or vice versa) (Figure \ref{fig:confusion_append} in the Appendix). This inconsistency resulted in a moderate correlation of only 0.3 between GPT-4’s satisfaction ratings and human ratings. This leaves uncertainties about the LLMs' abilities to capture the psychological states of the human players in negotiations.

%Future work on understanding the psychological personalities exhibited by LLMs can help to explain these observations.

% Confusion Matrix
\begin{figure}[bt]
    \centering
    \includegraphics[width=0.75\linewidth]{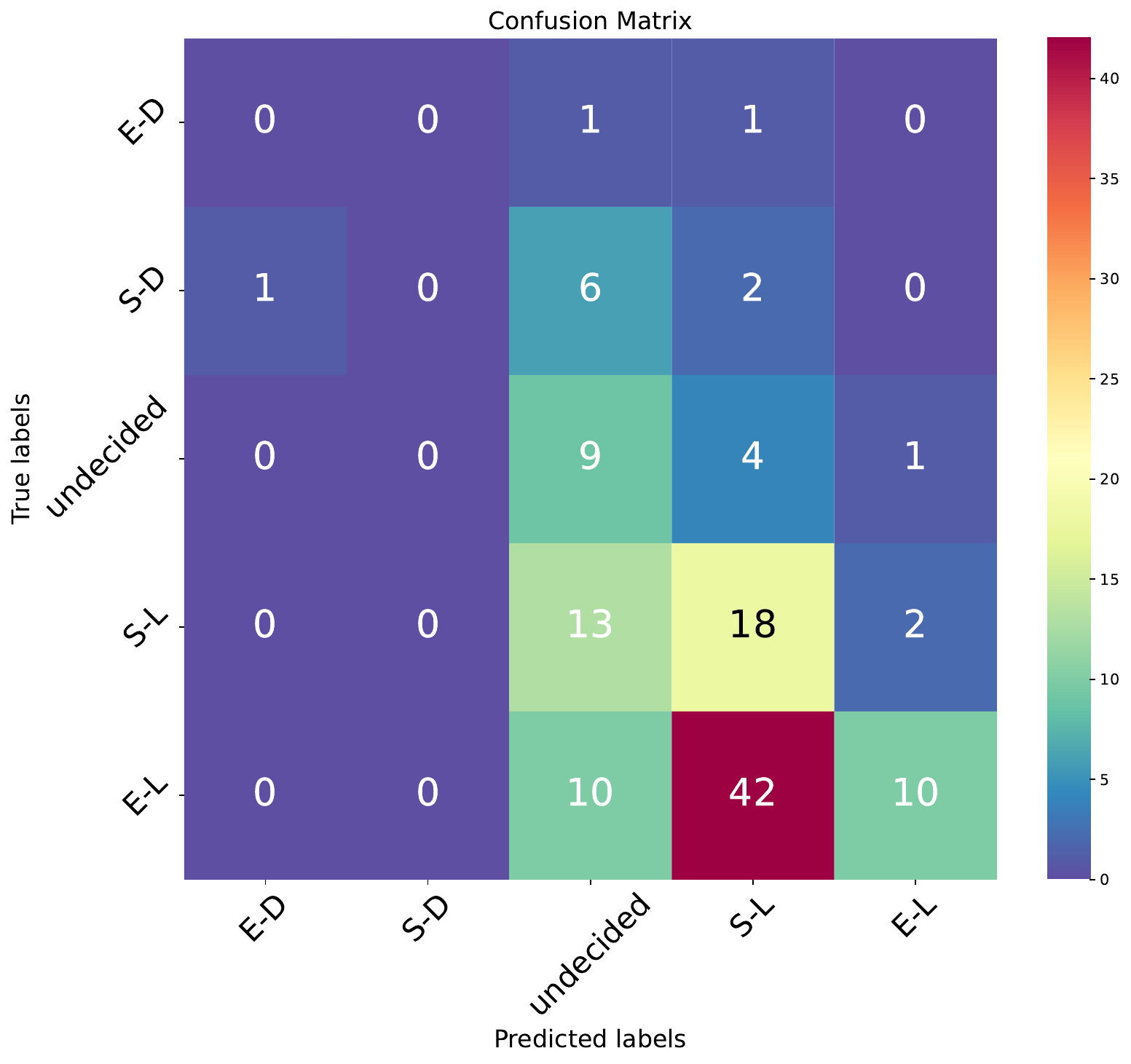}
    \caption{Confusion matrix of predictions of GPT-4 for the subjective task (\textit{end\_partner\_deal\_likeness\_ca}). \textit{E} stands for "Extremely", \textit{S} for "Slightly", \textit{D} for "Dislike" and \textit{L} for "Like."}
    \label{fig:confusion_subject}
\end{figure}

\subsection{Annotation Tasks}

Figure \ref{fig:main_d} shows that both GPT-3.5 and GPT-4 outperform the fine-tuned Flan-T5 baseline on annotation tasks, achieving Macro-F1 scores of 62.4\% and 52.5\%, respectively. Task-wise results (Table \ref{tab:full_results_all_models} in Appendix \ref{append:Details_of_Negotiation_Tasks}) indicate that these trends are influenced by Flan-T5’s poor performance on the \textit{dur\_dial\_act\_ji} task. However, Flan-T5 performs better than GPT-4 on most tasks, highlighting room for improvement in zero-shot scenarios. Additionally, models struggle more with detecting negotiation strategies than dialogue acts, likely due to the subtlety of strategy expressions. These results are without few-shot examples or prior utterances, which we will explore further in Section \ref{sec:discussion}.

\subsection{Partner Modeling Tasks}
Figure \ref{fig:main_e} presents the results for objective partner modeling tasks (inferring partner priorities) with the entire dialogue included in the prompt. GPT-4 achieves the best performance, demonstrating strong out-of-the-box Theory of Mind (ToM) abilities, unlike smaller open-source variants that mostly perform similarly to the Majority baseline. In Section \ref{sec:discussion}, we further analyze how model performance varies with the number of utterances seen by the model.

For inferring the partner’s subjective assessments (Figure \ref{fig:main_f}), Flan-T5 achieves an accuracy of 50.5\%, while GPT-4 scores the highest PCC at 0.39, highlighting the generally poor performance of models in assessing subjective perceptions. The skewed distribution in the confusion matrix (Figure \ref{fig:confusion_subject}) shows GPT-4 frequently predicts neutral or slight favorability in over 50\% of cases where partners demonstrate strong likeness, suggesting that GPT-4 fails to capture the degree of positivity that humans display. Prior work suggests that including partner demographics, personality, and emotional expression in the dialogue can improve these predictions~\cite{chawla2023towards}.

%(Figure \ref{fig:heatmap_b} in the Appendix) 

% CoT Prompting
\begin{figure}[bt]
    \centering
    \includegraphics[width=0.9\linewidth]{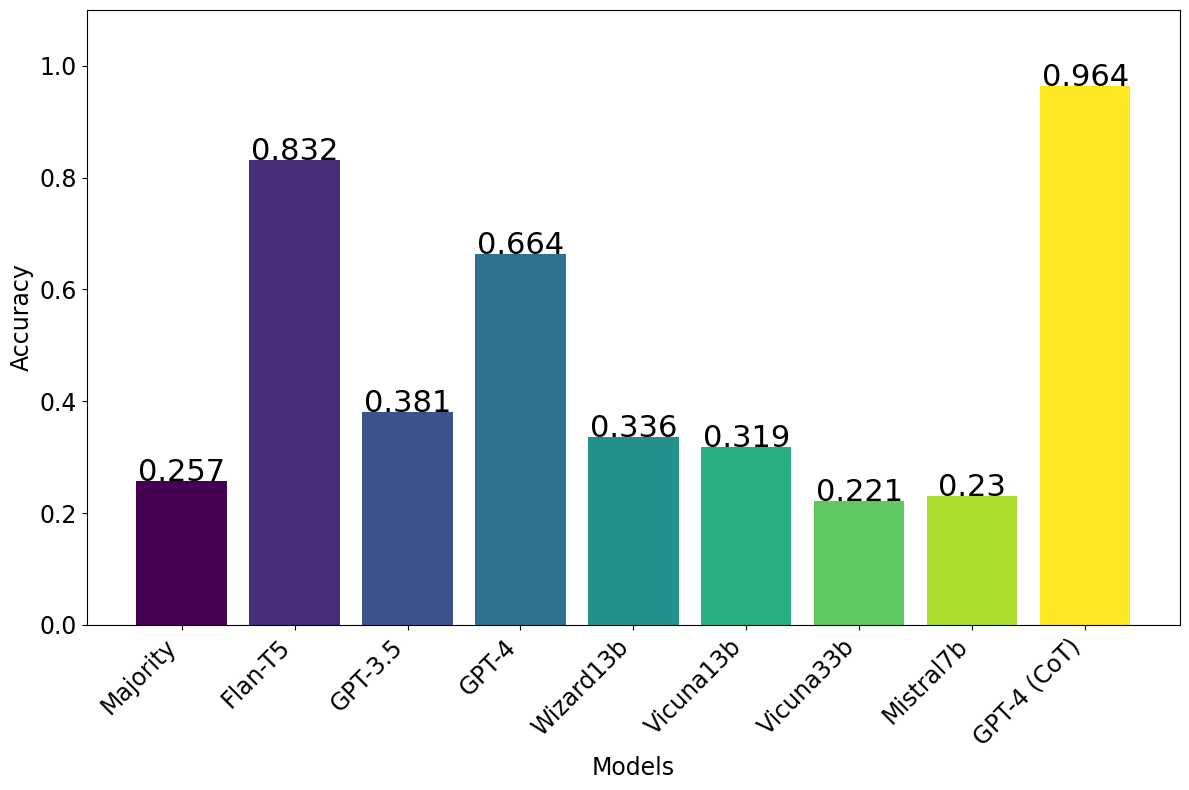}
    \caption{GPT-4’s evaluation on the \textit{end\_deal\_total\_dnd} task, highlighting the impact of Chain-of-Thought (CoT) prompting. Results for other tasks can be found in Figure \ref{fig:CoT_prompting_append} in the appendix.}
    \label{fig:CoT_prompting_main}
\end{figure}

\subsection{Generation Tasks}
\label{sec:generation-results}

Table \ref{tab:generation_results} shows that Flan-T5 significantly outperforms other models on BLEU and ROUGE, likely due to dataset-specific fine-tuning. LLMs struggle to align with the dataset’s utterance style and structure, leading to low overlap with reference (examples in Appendix \ref{sec:generation-samples-appendix}). However, on the CA dataset, which is richer and more diverse than DND, GPT models slightly surpass fine-tuned T5 in BERTScore, demonstrating better semantic understanding and generating responses semantically more similar to human reference.

Automatic evaluation has limitations in accurately assessing the appropriateness of model responses in complex negotiation scenarios. Therefore, we additionally conducted an expert human evaluation on a subset of 50 examples and confirmed a high inter-rater agreement with an Intra-Class Correlation Coefficient (ICC)~\cite{koo2016guideline} of 0.82. The human reference baseline performs the best on both \textit{Coherence} and \textit{Strategy} for both datasets. However, GPT-4 showed comparable \textit{Coherence} scores to humans, indicating it performs similarly to humans. In \textit{Strategy}, only the GPT-4 model approached human performance on the \textit{CA} dataset, but in all other cases, all models significantly underperformed compared to human references. Further, the models perform better on \textit{Coherence} than \textit{Strategy} -- generating a fluent and contextually appropriate response is easier than generating a strategic response aligned with the task goals. This can be attributed to the standard autoregressive training followed for these LLMs. While the models learn to output a plausible next utterance, they fail to do so strategically -- indicating task-specific tuning might still be necessary for improving performance in strategic social influence settings. Finally, the models perform much worse on \textit{Strategy} for \textit{CA} than \textit{DND}, with the former involving a more complex negotiation scenario and richer language use.

Although GPT-4’s performance is comparable to humans in our tasks, it is \textit{insufficient} for downstream applications in conversational AI and pedagogy. Prior work in negotiation agents~\cite{lewis2017deal, he2018decoupling, chawla2023selfish} relies on a combination of supervised learning (learning average behavior) and RL (explicit training for task-specific rewards). Therefore, we perform an error analysis and discuss key limitations to guide future work.

%%%%%%%%%%%%%%%%%%%%%%%%%%%%%%%%%%
\begin{figure}[tb]
    \centering
    % 첫 번째 subfigure
    \begin{subfigure}[b]{0.48\textwidth}  % width는 적절히 조정 가능
        \centering
        \includegraphics[width=0.9\textwidth]{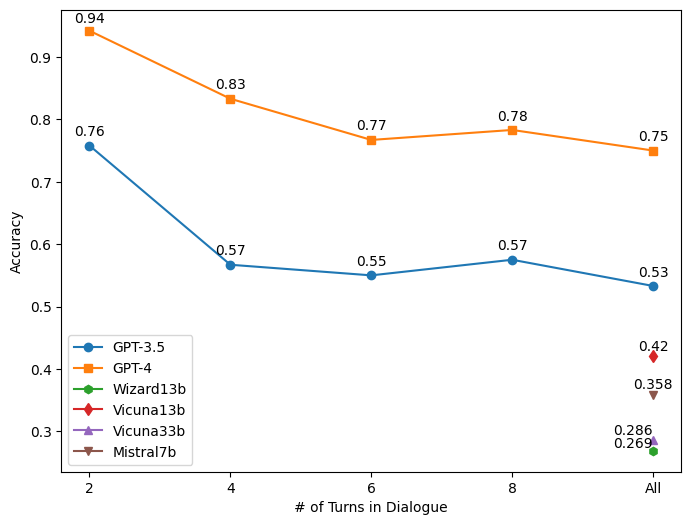}
        \caption{\textit{dur\_ask\_low\_priority\_ca}}
        \label{fig:dynamic_priority_a}
    \end{subfigure}
    
    % 두 번째 subfigure
    \begin{subfigure}[b]{0.48\textwidth}  % width는 적절히 조정 가능
        \centering
        \includegraphics[width=0.9\textwidth]{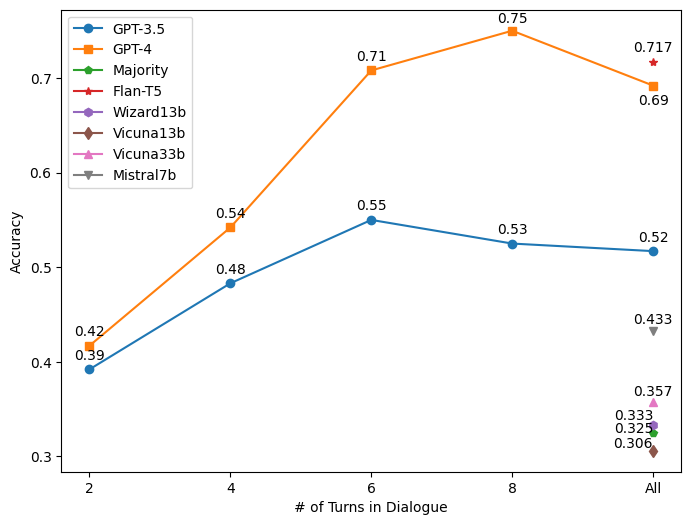}
        \caption{\textit{dur\_partner\_ask\_low\_priority\_ca}}
        \label{fig:dynamic_priority_b}
    \end{subfigure}
    
    \caption{Evaluation on the tasks demonstrating the impact of increasing the number of utterances seen by the models. }
    \label{fig:dynamic_priority}
\end{figure}
%%%%%%%%%%%%%%%%%%%%%%%%%%%%%%%%%%%%%
%%%%%%%%%%%%%%%%%%%%%%%%%%%%%%%%%%%%

\begin{table*}[hbt]
\centering
\resizebox{0.75\linewidth}{!}{%
\begin{tabular}{@{}ccccccccc@{}}
\multirow{2}{*}{Tasks}                                                               & \multirow{2}{*}{\begin{tabular}[c]{@{}c@{}}Label-Balance\\ (Tasks Difficulty)\end{tabular}} & \multirow{2}{*}{Metric} & \multicolumn{6}{c}{Model}                                       \\ \cmidrule(l){4-9} 
&                                                                                             &                         & GPT-3.5 & GPT-4 & Mistral7b & Vicuna13b & Vicuna33b & Wizard13b \\ \midrule
\multirow{8}{*}{\begin{tabular}[c]{@{}c@{}}High / Low Priority\\ Tasks\end{tabular}} & \multirow{4}{*}{\begin{tabular}[c]{@{}c@{}}Well-Balacned\\ (Easy)\end{tabular}}             & Acc. $\uparrow$                   & 0.677   & \textbf{0.91}  & 0.419     & 0.206     & 0.243     & 0.38      \\
&                                                                                             & F1 $\uparrow$                     & 0.669   & \textbf{0.908} & 0.328     & 0.14      & 0.115     & 0.348     \\
&                                                                                             & |Acc. - F1| $\downarrow$            & 0.007   & \textbf{0.002} & 0.091     & 0.066     & 0.128     & 0.032     \\
&                                                                                             & KL-D$\downarrow$                    & 0.189   & \textbf{0.02}  & 0.626     & 1.134     & 1.286     & 0.372     \\ \cmidrule(l){2-9} 
& \multirow{4}{*}{\begin{tabular}[c]{@{}c@{}}Well-Balacned\\ (Hard)\end{tabular}}             & Acc.$\uparrow$                    & 0.638   & \textbf{0.825} & 0.458     & 0.538     & 0.331     & 0.307     \\
&                                                                                             & F1$\uparrow$                      & 0.623   & \textbf{0.824} & 0.453     & 0.53      & 0.215     & 0.184     \\
&                                                                                             & |Acc. - F1| $\downarrow$             & 0.015   & \textbf{0.001} & 0.006     & 0.008     & 0.116     & 0.123     \\
&                                                                                             & KL-D$\downarrow$                    & 0.08    & \textbf{0.019} & 0.069     & 0.046     & 0.978     & 1.038     \\ \midrule
\multirow{4}{*}{\begin{tabular}[c]{@{}c@{}}Dial-Act / Strategy\\ Tasks\end{tabular}} & \multirow{4}{*}{\begin{tabular}[c]{@{}c@{}}Imbalanced\\ (Hard)\end{tabular}}                & Acc$\uparrow$                     & 0.853   & \textbf{0.898} & 0.742     & 0.71      & 0.81      & 0.641     \\
&                                                                                             & F1 $\uparrow$                    & 0.525   & \textbf{0.624} & 0.334     & 0.39      & 0.43      & 0.323     \\
&                                                                                             & |Acc - F1| $\downarrow$              & 0.328   & \textbf{0.274} & 0.408     & 0.32      & 0.38      & 0.318     \\
&                                                                                             & KL-D$\downarrow$                    & \textbf{0.732}   & 0.951 & 1.952     & 1.442     & 1.13      & 1.552    
\end{tabular}}
\caption{Evaluation results of LLMs on balanced and imbalanced tasks. The KL-D metric represents KL-Divergence. The reported values are average metrics of the tasks, selected based on the task type, label type (nominal class), label distribution, and generation validity (i.e., no null values), as described in the table \ref{tab:full_results_all_models} in the Appendix. Further details on the selected tasks can be found in Appendix ~\ref{append:LLM_evaluation_tasks}.}
\label{tab:KLD}
\end{table*}

%%%%%%%%%%%%%%%%%%%%%%%%%%%%%%%%%%%%
\noindent\textbf{Sophisticated \textit{Incoherent} Generations}: Handling off-topic interaction remains challenging, and trivial issues like repetition still persist. We also found sophisticated coherence errors. For example, given the dialogue history from \textit{DND}, ``\textit{...YOU: I can't accept that. how about I get the book and 1 hat, and you can get the balls and 1 hat?
THEM: I can't accept that deal... can I have one of each ?}'', Wizard13b replies: ``\textit{YOU: I understand you want one of each. How about I get the book and one hat, and you get the balls and one hat? That way, we both get something we want.}''. Although the output seems novel, it paraphrases YOU’s previous offer, which the partner already rejected. Ensuring coherence remains an open challenge for LLMs.

\noindent\textbf{Failure to Incorporate Strategic Cues}: Generating strategically sound responses is difficult for all models. The models often fail to cater to their own priorities and struggle to utilize the information shared by the user. In one case from \textit{CA}, even though the partner says, ``\textit{I am good at making fire and have a starter so I can give up all the wood.}'', the models fail to use this information in their counter offers.

\noindent\textbf{Overly Agreeable Behavior}: In general, the models tend to show agreement even with unfair offers, indicating that they are open for exploitation. We tried varying the prompts to adjust the model behavior, but this proved to be challenging. In the future, we suggest augmenting LLMs with task-specific RL policies for enhanced strategic reasoning.

\section{Discussion}
\label{sec:discussion}
We now analyze the impact of popular prompting strategies, focusing primarily on GPT-4, given its superior overall performance, and also evaluate prediction balance and robustness in LLMs.

\noindent\textbf{CoT Prompting}: We focus on tasks requiring multi-hop arithmetic reasoning, such as calculating the maximum possible points and inferring actual total points in a negotiation. Using CoT with GPT-4 yields nearly 100\% accuracy on the tasks we tested (Figure \ref{fig:CoT_prompting_main}), highlighting its effectiveness for LLMs in negotiation tasks. A sample CoT prompt is provided in Table \ref{tab:cot_prompt_example} in Appendix.

\noindent\textbf{Using Prior Utterances for Annotation Tasks}: We include two prior utterances as adding additional contexts in prompts for annotation tasks. The results are mixed (Figure \ref{fig:multi-shot-prompting} in Appendix \ref{append:LLM_evaluation_tasks})—performance improves in only one task and shows minor degradation in the other three. The type of annotation labels and data collection methodology likely influence these outcomes, and irrelevant utterances might confuse the models. Including prior utterances should be a domain-specific choice based on validation performance.

\noindent\textbf{Few-shot Prompting}: In-Context Learning (ICL) is a key ability of LLMs, enabling them to better understand tasks and follow instructions~\cite{wei2022emergent}. We examine whether adding two randomly sampled examples to annotation tasks enhances performance via ICL. Our findings show that ICL improves performance on two out of four tasks, indicating that few-shot examples can boost model performance (Figure \ref{fig:multi-shot-prompting} in Appendix \ref{append:LLM_evaluation_tasks}). Selecting optimal examples for ICL remains an active research area, which could benefit tasks involving strategic interactions.

\noindent\textbf{Varying the number of seen utterances} We explore two questions: 1) Do additional utterances confuse the model about its own context provided in the prompt? and 2) Does the model utilize additional utterances to infer the partner’s context?

The answer to both questions is `Yes’ (Figure \ref{fig:dynamic_priority}). When asked about its own preferences, model performance degrades with more utterances. However, performance improves when inferring partner preferences with additional utterances. This can be explained by \textit{recency bias}—models focus more on recent information, improving partner modeling but diminishing comprehension of their own context given at the start.

\subsection{Evaluating Prediction Balance and Robustness in LLMs}
We test LLMs’ ability to make balanced predictions to assess model robustness. As shown in Table \ref{tab:KLD}, GPT-4 performs best in both easy and hard well-balanced priority tasks, with the smallest differences in accuracy and F1 score, and near-zero KL-Divergence (KL-D), indicating highly balanced predictions. In imbalanced tasks, although all models show decreased F1 scores relative to accuracy, GPT-4 has the smallest decline and low KL-D, indicating relatively balanced predictions compared to other models. Our results confirm GPT-4’s superior robustness compared to other LLMs, as its balanced predictions on unseen data demonstrate enhanced generalization ability.

\section{Related Work}

% Negotiation Agents % Recent research has seen the development of negotiation agents using LLMs in diverse scenarios.
\noindent \textbf{Negotiation Agents}: \citet{lewis2017deal} pioneered the development of end-to-end negotiation dialogue systems using self-play Reinforcement Learning (RL).~\citet{chawla2023selfish} enhanced this work by employing tougher user simulators and utility-based rewards. Other efforts focused on dialogue-act-based RL policies for modular agents, such as for buyer-seller negotiations~\cite{he2018decoupling}. \citet{fu2023improving} designed an LLM-based agent for balloon price negotiations via self-play and feedback from a critic-LLM. \citet{gandhi2023strategic} used LLMs to improve strategic reasoning based on few-shot CoT prompting and demonstrations about states, values, and beliefs. \citet{abdelnabi2023llmdeliberation} focused on interactive multi-agent games, showing that agents can consistently reach successful deals through systematic zero-shot CoT prompting. While these studies employed LLMs for designing end-to-end agents, a systematic exploration of their diverse abilities is missing, inhibiting their use in negotiation-related use cases. Hence, we evaluate LLMs across various tasks to test their multifaceted abilities in negotiation interactions.

% like comprehension, reasoning, numeracy, and ToM
\noindent\textbf{Probing LLMs}: Numerous recent efforts focus on probing LLMs' abilities in a variety of domains~\cite{brown2020language, kosinski2023theory,kojima2023large,noever2023numeracy,ziems2023can}. LLMs tend to struggle in planning and solving complex mathematical, logical, and reasoning problems~\cite{hao2023reasoning, huang2023reasoning}. While they show promise in ToM tasks \cite{bubeck2023sparks, kosinski2023theory}, \citet{kosinski2023theory} argues that their understanding is superficial, and \citet{ullman2023large} provides evidence for their brittle performance. We contribute to this line of work by specifically probing LLMs in complex negotiation scenarios, covering a range of tasks from those requiring one-step reasoning to response generation that requires a blend of conversational understanding, inference about the other party's needs, and reasoning for strategic decision-making.

\section{Conclusion}
\label{sec:conclusion}

We devise a methodology to systematically analyze the multifaceted capabilities of LLMs in negotiations. When evaluated out-of-the-box, GPT-4 outperforms and is more robust than other LLMs but still leaves room for improvement in most tasks. CoT and few-shot prompting help improve performance in several arithmetic reasoning and annotation tasks. In contrast, smaller open-source models struggle, performing comparably to the trivial Majority baseline for key Comprehension and Partner Modeling tasks.

Based on our results, we conclude that LLMs can indeed be helpful across a number of use cases in negotiation research. This is not only limited to designing dialogue systems but also includes scaling-up pedagogical and data collection practices. LLMs capable of partner modeling can help to provide feedback to students who fail to elicit and incorporate their partner's preferences~\cite{johnson2019intelligent}. Our results on annotation tasks indicate that LLMs can make the annotation process efficient (albeit with a human in the loop), aiding both linguistic strategy analysis and the design of modular dialogue systems. However, this is primarily true for proprietary LLMs, emphasizing the need to improve smaller open-source models.

% We also find that extra information in the dialogue can confuse the models about their own context, provided explicitly in the prompts. Future work must develop ways to enable LLMs to handle longer contexts better, focusing only on relevant parts of the input. Our human evaluation shows that all models struggle with generating strategically appropriate responses. Recent efforts in structured reasoning with LLMs can help mitigate this issue~\cite{zhou2024self}, as these approaches are suited for strategic dialogue. Response generation can be broken down into subgoals, such as sequentially performing comprehension, annotation, and partner modeling before outputting the final response. In the future, we plan to combine LLMs with RL policies explored in prior negotiation work~\cite{he2018decoupling} to control the strategy or personality of the dialogue agent.

We also find that information in dialogue can confuse models about their own context, even when provided in the prompts. Future work should focus on helping LLMs handle longer contexts by emphasizing relevant input. Human evaluation shows all models struggle with generating strategically appropriate responses. Recent efforts in structured reasoning~\cite{zhou2024self} may address this by breaking response generation into subgoals like comprehension, annotation, and partner modeling. We plan to combine LLMs with RL policies from prior negotiation work~\cite{he2018decoupling} to control the dialogue agent’s strategy or personality.

\section{Broader Impact and Ethical Considerations}

\subsection{Datasets}
Our study used four publicly available negotiation datasets (i.e., CRA, DND, CA, and JI), which were thoroughly anonymized before their release by the respective authors. We conducted a meticulous review of the licensing details for each dataset to ensure that our usage strictly adheres to their intended purposes and scope. We note that all datasets are in English, so it is unclear if the same findings extend to other languages or cultures. In fact, differences in how people negotiate across cultures have received significant attention in the literature~\cite{luo2008analysis,andersen2018cultural}, and thus, we encourage future work to investigate LLM negotiation capabilities in other languages as well. Our methodology for designing tasks and evaluation procedures is language-independent, and we hope that it can guide future efforts in this direction.

\subsection{LLMs}
We used LLMs strictly within the intended scope in accordance with the respective licensing details. Our approach is consistent with various other recent efforts that aim to evaluate the diverse capabilities of LLMs, ensuring that the use remains within ethical and operational guidelines. 

%In addition, we meticulously adhered to all licensing agreements associated with the LLMs to ensure their usage was both legally and ethically compliant.

% Furthermore, potential ethical issues such as data privacy and various biases that may arise from the output generated by the LLM, were addressed through a thorough review of the output and conducting bias assessments.

\subsection{Human Evaluation}

We gathered expert human annotations to evaluate the responses generated by our compared models. The evaluation of negotiation dialogues requires domain knowledge and sophisticated assessment, hence it was conducted by five expert annotators, including the authors of this work who possess expertise in this field. General crowdsourcing was not ideal for this evaluation.

\subsection{AI for Social Influence Interactions}
Negotiation dialogues fall under the broader spectrum of social influence tasks~\cite{chawla-etal-2023-social}, which target achieving specific changes in behaviors or opinions through conversations (other example scenarios include online toxicity moderation, therapy, argumentation, etc.). Automated systems that can comprehend or participate in such interactions find broad applications in conversational AI and pedagogy through the development of tools that can make everyday social interactions more effective and efficient.

\noindent\textbf{Ethical Recommendations}: Naturally, as for any human-facing technology, efforts in this area also raise ethical concerns that must be properly addressed. This includes possibilities for manipulation, potential misuse, bias, and discrimination~\cite{lewicki2016essentials}.

We provide four key recommendations here: 1) Maintaining \textit{\textbf{transparency}} about the dataset and model design processes, along with the known capabilities and misbehaviors of the developed systems, 2) Ensuring proper \textbf{\textit{consent}} procedures, 3) Continuous \textbf{\textit{monitoring}} of the designed systems, and 4) Using forums like the *ACL Workshop Series on Social Influence in Conversations (SICon)\footnote{\url{https://sites.google.com/view/sicon-2023/home}} for a \textbf{\textit{principled discussion}} on this topic.

\section{Limitations}

\noindent\textbf{Task Design}: The datasets used in our analysis are based on a framework from the negotiation literature, referred to as the Multi-Issue Bargaining Task or MIBT~\cite{fershtman1990importance}. MIBT has been a popular framework for defining negotiation scenarios, both in academic and industrial settings. However, being an abstraction of real-world negotiations, it misses out on several real-world aspects, such as when the player preferences change (i.e., dynamic change) during the interaction or when individual items can be broken down into subparts for fractional division between the players. We encourage future work to take up these other complex scenarios as well.

\noindent\textbf{Prompting Variations}: We primarily evaluated LLMs using zero-shot prompts to test out-of-the-box capabilities. We explored CoT and few-shot prompting for a subset of tasks to gain additional insights. Although we designed the prompts based on careful experimentation and consideration following the best practices from prior work, we acknowledge that other ways of prompting the models with more sophisticated prompt engineering methods could potentially lead to different results. This is an active area of research. While our goal in this work was to cover the breadth of capabilities based on standard prompting techniques, we encourage future work to investigate the impact of prompt engineering in-depth, albeit on a smaller number of tasks.

\section{Acknowledgments}
Research was sponsored by the Army Research Office under Cooperative Agreement Number W911NF-20-2-0053. The views and conclusions contained in this document are those of the authors and should not be interpreted as representing the official policies, either expressed or implied, of the Army Research Office or the U.S. Government. The U.S. Government is authorized to reproduce and distribute reprints for Government purposes, notwithstanding any copyright notation herein.

\bibliography{anthology,custom}

\clearpage
\appendix

\section{Details of Negotiation Tasks}
\label{append:Details_of_Negotiation_Tasks}
We present an overview and detailed task descriptions of all the designed tasks in Table \ref{tab:details_of_all_tasks} and Table \ref{tab:task_desc_of_all_tasks}. Table \ref{tab:dist_of_tasks} summarizes the distribution of tasks across datasets and time stages. In some cases, a specific model failed to follow the provided instructions for a specific task and hence, failed to produce reliable results. Such tasks were entirely excluded (for all models) while computing the average bar plots in Figure \ref{fig:result_task_type} for a fair comparison. Table \ref{tab:selected_tasks} presents the final list of selected tasks used for computing these plots. Finally, we report the task-wise results for all models in Table \ref{tab:full_results_all_models} for completeness.

% Meanwhile, Tables \ref{tab:selected_tasks} and \ref{tab:full_results_all_models} contain information on the tasks selected for calculating the evaluation results of the LLM by Task Type and the evaluation results of the LLMs for all tasks, respectively.

\section{Baselines}
\label{append:details_bs_training}
% \subsection{Baseline model}
We use two baselines in this work: a \textit{Majority} baseline, that trivially outputs the majority answer from the training data, and a \textit{Flan-T5} fine-tuning baseline, where we train a model for each task separately by fine-tuning on the available training data.

\noindent\textbf{Task Selection}: For all tasks that are defined in the \textit{Start} time stage, it is possible to achieve a $100\%$ accuracy through trivial rule-based parsers. Hence, we excluded these tasks for evaluation with \textit{Majority} and \textit{Flan-T5}. 

% Given the nature of tasks from the start stage, where it is possible to achieve 100\% accuracy through simple rule-based parsing or basic calculation, it was deemed inappropriate as tasks for model training. Consequently, for selecting tasks to evaluate the Majority-Label model and the FT-5 model, \textit{Start} tasks were excluded from the baseline evaluation.

\noindent\textbf{Data Preprocessing}: During the data preprocessing for fine-tuning, although there are differences in the information included in the prompts for each task, prompts for evaluating LLMs are generally lengthy due to detailed task descriptions. To enhance training efficiency during fine-tuning, we removed these lengthy descriptions from the inputs to the Flan-T5 model and instead replaced them with simple task instructions such as `predict highest priority', similar to using brief descriptions when typically pretraining Flan-T5 in a multi-task setting.

After applying the previously described preprocessing methods, we constructed the final dataset and, excluding 200 test cases for LLM evaluation, split the remaining data into training and development sets in a 9:1 ratio for model training and evaluation.
% This decision was based on the assessment that the removed descriptions are not directly relevant to the model's prediction accuracy and their omission would not negatively impact the learning process. Instead of the removed task descriptions, we inserted simple task instructions such as "predict highest priority", similar to using brief task descriptions as input when typically pretraining T5 in a multi-task setting.

\noindent \textbf{Multi-task Training}
In line with how Flan-T5 was originally trained, we conducted our experiments in two ways: 1) Training a single model on all tasks together (FT-5-All-Task) and 2) Training one model for each individual task separately (FT-5-By-Task). The results showed that the FT-5-All-Task model generally underperformed compared to the FT-5-By-Task models, with observed poor learning in several tasks. Consequently, FT-5-By-Task models were chosen as a baseline.

%dataset preprocessing details
%Task 뺀거
% All task vs by task
% \subsection{Details of Computation}
\noindent\textbf{Compute Resources}: We trained the baseline model (Flan-T5) for each task under various experimental settings and hyperparameter adjustments, utilizing over 500 hours of GPUs such as NVIDIA V100 and A100 GPUs. 

\noindent\textbf{Implementation}: For fine-tuning Flan-T5, we used the released model\footnote{https://huggingface.co/google/flan-t5-base} on the Hugging Face model hub as a back-bone model. The evaluation code for model assessment was developed from scratch, while for evaluating generated responses, we utilized existing packages for BLEU\footnote{https://pypi.org/project/evaluate/} and ROUGE-L\footnote{https://pypi.org/project/rouge-score/} respectively. %The evaluation code for baseline  were utilized in evaluating LLMs in the same manner.
% \onecolumn
% \clearpage

\section{Statistical Power Analysis for Test set}\label{append:power_analysis}
We used a capped test set of 200 samples across tasks. To ensure this sample size was sufficient to detect model differences, we conducted a Chi-squared power analysis. With a significance level of p = 0.05, a medium effect size (W = 0.3), and four outcome categories, the power for 200 samples was 0.959, confirming adequacy for detecting meaningful differences.

We applied the McNemar test to assess model performance~\cite{demvsar2006statistical}, focusing on T5 and GPT-4 across all tasks with valid results (excluding the subjective tasks with ordinal variables) in Table ~\ref{tab:full_results_all_models}. As shown in Table ~\ref{tab:power_test} below, when performance was similar, no significant differences were found, but statistically significant differences were detected where the models’ outcomes diverged. This confirmed that the sample size was sufficient to detect differences between models.

\begin{table}[H]
\centering
\resizebox{\linewidth}{!}{%
\begin{tabular}{@{}lrrrr@{}}
\toprule
\multicolumn{1}{c}{\multirow{2}{*}{Task name}} & \multicolumn{2}{c}{Models}                         & \multicolumn{2}{c}{McNeMar's test}                                                                             \\ \cmidrule(l){2-5} 
\multicolumn{1}{c}{}                           & \multicolumn{1}{c}{T5} & \multicolumn{1}{c}{GPT-4} & \multicolumn{1}{c}{\begin{tabular}[c]{@{}c@{}}Chi-square \\ (DF:1)\end{tabular}} & \multicolumn{1}{c}{P-value} \\ \midrule
dur\_dial\_act\_cra                            & 0.787                  & 0.678                     & 15.63                                                                            & $ 0.0001^{*}  $                    \\
dur\_dial\_act\_dnd                            & 0.96                   & 0.825                     & 5.14                                                                             & $0.0233^{*}  $                   \\
dur\_dial\_act\_ji                             & 0.019                  & 0.578                     & 145.31                                                                           & $0.0001^{*} $                    \\
dur\_full\_proposal\_cra                       & 0.439                  & 0.369                     & 2.95                                                                             & 0.0859                      \\
dur\_full\_proposal\_dnd                       & 1                      & 0.866                     & 79.01                                                                            & $0.0001^{*} $                     \\
dur\_partner\_ask\_high\_priority\_ca          & 0.717                  & 0.792                     & 1.49                                                                             & 0.2225                      \\
dur\_partner\_ask\_low\_priority\_ca           & 0.717                  & 0.75                      & 0.1                                                                              & 0.7488                      \\
dur\_strategy\_ca                              & 0.724                  & 0.507                     & 4.97                                                                             & $0.0259^{*}$                      \\
end\_deal\_specifics\_ca                       & 0.364                  & 0.664                     & 177.03                                                                           & $0.0001^{*}$                      \\
end\_deal\_specifics\_dnd                      & 0.973                  & 0.67                      & 40.45                                                                            & $0.0001^{*}$                      \\
end\_deal\_specifics\_ji                       & 0.764                  & 0.858                     & 73.29                                                                            & $0.0001^{*} $                     \\
end\_deal\_total\_ca                           & 0.233                  & 0.083                     & 9.63                                                                             & $0.0019^{*}$                      \\
end\_deal\_total\_dnd                          & 0.832                  & 0.664                     & 9.26                                                                             & $0.0023^{*}  $                    \\ \bottomrule
\end{tabular}}
\caption{McNeMar’s test results for the two models. * indicates statistical significance (P-value < 0.05)
 }
\label{tab:power_test}
\end{table}

\begin{table*}[ht]
\centering
\resizebox{0.6\linewidth}{!}{%
\begin{tabular}{cccc}
Dataset & Time Stage & Full Task Name                             & Task Type\\ \midrule
CA      & Start             & sta\_ask\_high\_priority\_ca          & Comprehension\\
CA      & Start             & sta\_ask\_low\_priority\_ca           & Comprehension\\
CA      & Start             & sta\_ask\_point\_values\_ca           & Comprehension\\
CA      & Start             & sta\_max\_points\_ca                  & Comprehension\\
CA      & Start             & sta\_total\_item\_count\_ca           & Comprehension\\
CA      & During            & dur\_partner\_ask\_high\_priority\_ca & Partner Modeling\\
CA      & During            & dur\_partner\_ask\_low\_priority\_ca  & Partner Modeling\\
\end{tabular}}
\caption{An overview of the designed tasks for evaluating LLMs in negotiations. \textit{CRA}: \cite{devault2015toward}, \textit{DND}: \cite{lewis2017deal}, \textit{CA}: \cite{chawla-etal-2021-casino}, \textit{JI}: \cite{yamaguchi-etal-2021-dialogue}.}
\label{tab:details_of_all_tasks}
\end{table*}

\begin{table*}[ht]
\centering
\resizebox{0.80\linewidth}{!}{%
%
}
\caption{Task descriptions of all designed tasks for evaluating LLMs}
\label{tab:task_desc_of_all_tasks}
\end{table*}

\begin{table*}[ht]
\centering
\resizebox{0.3\linewidth}{!}{%
\begin{tabular}{@{}lllll@{}}
\multirow{2}{*}{Dataset} & \multicolumn{3}{c}{Negotiation Stage} & \multirow{2}{*}{Total} \\ \cmidrule(lr){2-4}
                         & Start        & During        & End       &                        \\ \midrule
CA                       & 5            & 6          & 6         & 17                     \\
CRA                      &              & 2          &           & 2                      \\
DND                      & 3            & 3          & 2         & 8                      \\
JI                       & 2            & 5          & 1         & 8                      \\ \midrule
Total                    & 10           & 16         & 9         & 35                     \\
\end{tabular}}
\caption{Distribution of the designed tasks by the dataset and time stage.}
\label{tab:dist_of_tasks}
\end{table*}

\begin{table*}[ht]
\centering
\resizebox{\linewidth}{!}{%
%
}
\caption{Selected tasks for computing the average bar plots in Figure \ref{fig:result_task_type}.}
\label{tab:selected_tasks}
\end{table*}

\begin{table*}[ht]
\centering
\resizebox{\linewidth}{!}{%
\begin{tabular}{@{}llllllllll@{}}
\multicolumn{1}{c}{\multirow{2}{*}{Full Task Name}} & \multicolumn{1}{c}{\multirow{2}{*}{Metric}} & \multicolumn{8}{c}{Model}                                                                                                                                                                                                                           \\ \cmidrule(l){3-10} 
\multicolumn{1}{c}{}                           & \multicolumn{1}{c}{}                             & \multicolumn{1}{c}{Majority} & \multicolumn{1}{c}{Flan-T5} & \multicolumn{1}{c}{GPT-3.5} & \multicolumn{1}{c}{GPT-4} & \multicolumn{1}{c}{Mistral7b} & \multicolumn{1}{c}{Vicuna13b} & \multicolumn{1}{c}{Vicuna33b} & \multicolumn{1}{c}{Wizard13b} \\ \midrule
end\_deal\_likeness\_ca                        & Acc./PCC                                         & 0.525/0                        & 0.525/0                       & 0.357/0.419                       & 0.175/0.367                     &                               & 0.119/-0.033                         & 0.267/0.245                         & 0.239/0.234                         \\
end\_deal\_satisfaction\_ca                    & Acc./PCC                                         & 0.5/0                          & 0.467/-0.008                       & 0.373/0.211                       & 0.417/0.304                     & 0.092/0.111                         & 0.266/0.001                         & 0.216/0.114                         & 0.445/0.118                         \\
end\_deal\_specifics\_ca                       & Acc.                                & 0.356                        & 0.364                       & 0.664                       & 0.916                     & 0.517                         & 0.517                         & 0.593                         & 0.555                         \\
\end{tabular}}
\caption{Task-wise results for all models. Empty values for Majority and Flan-T5 correspond to the tasks on which these baselines were not evaluated (since a simple rule-based baseline can achieve $100\%$ performance in these cases). Empty values for other LLMs indicate that the model failed to produce the intended or valid output for the given task-specific prompt, not passing our generation validity check (threshold: 80\% valid response rate), making the measured scores too unreliable to report. Notes: 1) PCC: Pearson Correlation Coefficient, 2) The results on response generation are provided in Table \ref{tab:generation_results}, and 3) For the cases where the models guess the outputs for each issue (like books, balls, or hats in DND) separately, we simply report the average score across all issues. The results for two generation tasks (i.e., \textit{dur\_gen\_resp\_ca}, \textit{dur\_gen\_resp\_dnd}) are excluded from the table and can be seen in Table \ref{tab:generation_results}.}
\label{tab:full_results_all_models}
\end{table*}

% \clearpage
% \onecolumn
\section{Prompting}
\label{append:prompting}
The precise versions used for GPT-3.5 and GPT-4 are `gpt-3.5-turbo-1106' and `gpt-4-1106-preview', respectively. Tables \ref{tab:prompt_sta_ask_point_values_ca}, \ref{tab:prompt_mid_full_proposal_dnd}, and \ref{tab:prompt_end_deal_specifics_ca} provide examples of prompts for specific tasks corresponding to the \textit{start}, \textit{during}, and \textit{end} time stages respectively. Additionally, Table \ref{tab:cot_prompt_example} showcases an example of CoT prompting used for the \textit{end\_deal\_total\_ca} task. Table \ref{tab:Questions_prompt} shows the questions used in prompts for all tasks.
\begin{table*}[ht]
\centering
\resizebox{\linewidth}{!}{%
\begin{tabular}{@{}l@{}}
Prompt example (Task: sta\_ask\_point\_values\_ca)\\ \midrule
\begin{tabular}[c]{@{}l@{}}\textcolor{blue}{Task Description}: You are negotiating with your campsite neighbor over an extra supply of food, water, and firewood for your camping trip. \\ Different types of packages are worth different amounts of points to each one of you. You'll be provided with information about the negotiation. \\ Then, you'll answer a question. \\ \\ Here are the number of food, water, and firewood packages available in the negotiation, contained in \textless{}count\textgreater tags. \\ \textcolor{blue}{\textless{}count\textgreater} Food Packages: 3 Water Packages: 3 Firewood Packages: 3 \textcolor{blue}{\textless{}/count\textgreater} \\ \\ Here are the number of points you get for each type of package, contained in \textless{}value\textgreater tags. \\\textcolor{blue}{\textless{}value\textgreater} Each Food Package: 3 points Each Water Package: 5 points Each Firewood Package: 4 points \textcolor{blue}{\textless{}/value\textgreater} \\ \\ \textcolor{blue}{Question}: How many points is one package of each issue worth to you? Present your answer as a json within \textless{}answer\textgreater \textless{}/answer\textgreater tags with \\ keys as issues (food, water, and firewood) and values as the corresponding answers.\end{tabular} \\
\end{tabular}}
\caption{Prompt example of the Start-stage Task: \textit{sta\_ask\_point\_values\_ca} task.}
\label{tab:prompt_sta_ask_point_values_ca}
\end{table*}

\begin{table*}[ht]
\centering
\resizebox{\linewidth}{!}{%
\begin{tabular}{@{}l@{}}
Prompt example (Task: dur\_full\_proposal\_dnd)                                  \\ \midrule
\begin{tabular}[c]{@{}l@{}}\textcolor{blue}{Task Description}: You are negotiating with a partner over some quantity of books, hats, and balls to determine who gets which items. \\ Different types of items are worth different amount of points to each one of you. You'll be provided with information about the negotiation. \\ 
Then, you'll answer a question. \\ \\ Here are the number of books, hats, and balls available in the negotiation, contained in \textless{}count\textgreater tags. \\ \textcolor{blue}{\textless{}count\textgreater} Books: 3 Hats: 1 Balls: 2 \textcolor{blue}{\textless{}/count\textgreater} \\ \\ Here are the number of points you get for each type of item, contained in \textless{}value\textgreater tags. \\ \textcolor{blue}{\textless{}value\textgreater} Each Book: 1 points Each Hat: 5 points Each Ball: 1 points \textcolor{blue}{\textless{}/value\textgreater} \\ \\ Here is an utterance from the negotiation, contained in \textless{}utterance\textgreater tags. \\ \textcolor{blue}{\textless{}utterance\textgreater} YOU: i'll take the hat and balls if you want the books \textcolor{blue}{\textless{}/utterance\textgreater} \\ \\ \textcolor{blue}{Question}: How many items does the speaker get for each issue in the proposal delimited by the \textless{}utterance\textgreater tags? Present your answer as a json \\ within \textless{}answer\textgreater \textless{}/answer\textgreater tags with keys as issues (books, hats, and balls) and values as the corresponding answers. If the answer is not clear \\ for an issue, output NA.\end{tabular} \\
\end{tabular}}
\caption{Prompt example of the During-stage Task: \textit{dur\_full\_proposal\_dnd} task.}
\label{tab:prompt_mid_full_proposal_dnd}
\end{table*}

\begin{table*}[ht]
\centering
\resizebox{\linewidth}{!}{%
\begin{tabular}{@{}l@{}}
Prompt example (Task: end\_deal\_specifics\_ca)                                  \\ \midrule
\begin{tabular}[c]{@{}l@{}}\textcolor{blue}{Task Description}: You are negotiating with your campsite neighbor over extra supply of food, water, and firewood for your camping trip. Different types of packages \\ are worth different amount of points to each one of you. You'll be provided with information about the negotiation. Then, you'll answer a question. \\ \\ Here are the number of food, water, and firewood packages available in the negotiation, contained in \textless{}count\textgreater tags. \\ \textcolor{blue}{\textless{}count\textgreater} Food Packages: 3 Water Packages: 3 Firewood Packages: 3 \textcolor{blue}{\textless{}/count\textgreater} \\ \\ Here are the number of points you get for each type of package, contained in \textless{}value\textgreater tags. \\ \textcolor{blue}{\textless{}value\textgreater} Each Food Package: 3 points Each Water Package: 5 points Each Firewood Package: 4 points \textcolor{blue}{\textless{}/value\textgreater} \\ \\ Here is the complete dialogue, contained in \textless{}dialogue\textgreater tags. \\ \textcolor{blue}{\textless{}dialogue\textgreater} THEM: Hello, I would like to have three packages of food. We've decided to stay an extra night but need more food to do so. \\YOU: I would be open to that if you could give me three packages of water \smiley{} \\THEM: Hmmm...I'm pretty muddy due to clumsiness, so I may need one extra. I could give you two waters and all of the firewood. What do you think? \smiley{} \\YOU: So are you suggesting that I would get 2 waters, 3 firewood, and no food? \\THEM: Right! Well, beyond the food you already have. \\YOU: I have an extra person camping with us that I didn't expect when I bought food, so I could use one if you're willing \smiley{} \\THEM: I understand that! I wasn't expecting to stay an extra night, but the weather is too perfect to leave. I can manage with two packages of food for sure. \smiley{} \\YOU: Great! Thank you for being so understanding! \\THEM: No problem! So are we in agreement that I get 2 food, 1 water and you get the reverse? I could also probably use one firewood, but it's not as important to me. \\YOU: I can give you one firewood, so I'll be getting 1 food, 2 water, and 2 firewood? \textcolor{blue}{\textless{}/dialogue\textgreater}\\ \\ \textcolor{blue}{Question}: In the final deal, how many item of each issue did you get? Present your answer as a json within \textless{}answer\textgreater \textless{}/answer\textgreater tags with keys as issues (food, water, \\and firewood)  and values as the corresponding answers. If there was no agreement, answer NA for each issue.\end{tabular} \\
\end{tabular}}
\caption{Prompt example of the End-stage Task: the end\_deal\_specifics\_ca task.}
\label{tab:prompt_end_deal_specifics_ca}
\end{table*}

\begin{table*}[ht]
\centering
\resizebox{\linewidth}{!}{%
\begin{tabular}{@{}l@{}}
Prompt example (Task: end\_deal\_total\_ca)                                         \\ \midrule
\begin{tabular}[c]{@{}l@{}}Task Description: You are negotiating with your campsite neighbor over extra supply of food, water, and firewood for your camping trip. Different types of packages \\ are worth different amount of points to each one of you. You'll be provided with information about the negotiation. Then, you'll answer a question.\\ \\ Here are the number of food, water, and firewood packages available in the negotiation, contained in <count> tags.\\ <count>\\ Food Packages: 3\\ Water Packages: 3\\ Firewood Packages: 3\\ </count>\\ \\ Here are the number of points you get for each type of package, contained in <value> tags.\\ <value>\\ Each Food Package: 3 points\\ Each Water Package: 5 points\\ Each Firewood Package: 4 points\\ </value>\\ \\ Here is the complete dialogue, contained in <dialogue> tags.\\ <dialogue>\\ THEM: Hello, I would like to have three packages of food. We've decided to stay an extra night but need more food to do so.\\ YOU: I would be open to that if you could give me three packages of water\\ THEM: Hmmm...I'm pretty muddy due to clumsiness, so I may need one extra. I could give you two waters and all of the firewood. What do you think? \\YOU: So are you suggesting that I would get 2 waters, 3 firewood, and no food?\\ THEM: Right! Well, beyond the food you already have.\\ YOU: I have an extra person camping with us that I didn't expect when I bought food, so I could use one if you're willing\\THEM: I understand that! I wasn't expecting to stay an extra night, but the weather is too perfect to leave. I can manage with two packages of food for sure.\\ YOU: Great!  Thank you for being so understanding!\\ THEM: No problem! So are we in agreement that I get 2 food, 1 water and you get the reverse? I could also probably use one firewood, but it's not as important to me.\\ YOU: I can give you one firewood, so I'll be getting 1 food, 2 water, and 2 firewood?\\ </dialogue>\\ \\ Question: How many points did you get at the end of the negotiation?\\ \\ NOTE: Let's think step-by-step! Put your thoughts in <thinking> </thinking> tags, and put your answer as a single number in <answer> </answer> tags.\end{tabular} \\
\end{tabular}}
\caption{CoT prompt example for the \textit{end\_deal\_total\_ca} task.}
\label{tab:cot_prompt_example}
\end{table*}

\begin{table*}[ht]
\centering
\resizebox{\linewidth}{!}{%
\begin{tabular}{@{}c|l@{}}
Task                                           & \multicolumn{1}{c}{Question}                                                                                                                                                                                                                                                                                                                                                                                                     \\ \midrule
\makecell[c]{\textit{sta\_total\_item\_count\_dnd}\\\textit{sta\_total\_item\_count\_ca}}         & What is the total number of items being negotiated over? Present your answer as a single number with no additional text.                                                                                                                                                                                                                                                                                                          \\ \midrule
% \textit{sta\_total\_item\_count\_ca}           & What is the total number of items being negotiated over? Present youranswer as a single number with no additional text.                                                                                                                                                                                                                                                                                                          \\ \midrule
\makecell[c]{\textit{sta\_max\_points\_dnd}\\\textit{sta\_max\_points\_ca}}                & What is the maximum number of points that you can possibly get in any deal? Present your answer as a single number with no additional text.                                                                                                                                                                                                                                                                                       \\ \midrule
% \textit{sta\_max\_points\_ca}                  & What is the maximum number of points that you can possibly get in anydeal? Present your answer as a single number with no additional text.                                                                                                                                                                                                                                                                                       \\ \midrule
\textit{sta\_ask\_point\_values\_dnd}          & \begin{tabular}[c]{@{}l@{}}How many points is one item of each issue worth to you? Present your answer as a JSON within \textless{}answer\textgreater \textless{}/answer\textgreater tags \\ with keys as issues (books, hats, and balls) and values as the corresponding answers.\end{tabular}                                                                                                                                  \\ \midrule
\textit{sta\_ask\_point\_values\_ca}           & \begin{tabular}[c]{@{}l@{}}How many points is one package of each issue worth to you? Present your answer as a JSON within \textless{}answer\textgreater \textless{}/answer\textgreater tags \\ with keys as issues (food, water, and firewood) and values as the corresponding answers.\end{tabular}                                                                                                                            \\ \midrule
\makecell[c]{\textit{sta\_ask\_low\_priority\_ji}\\\textit{dur\_ask\_low\_priority\_ji}}           & \begin{tabular}[c]{@{}l@{}}What is your lowest priority issue? Present your answer as one of the following multiple choice options. You must select an option.\\  A: position / B: company / C: salary / D: days\_off / E: workplace\end{tabular}                                                                                                                                                                                \\ \midrule
\makecell[c]{\textit{sta\_ask\_low\_priority\_ca}\\\textit{dur\_ask\_low\_priority\_ca}}           & \begin{tabular}[c]{@{}l@{}}What is your lowest priority issue? Present your answer as one of the following multiple choice options. You must select an option.\\  A: food / B: water / C: firewood\end{tabular}                                                                                                                                                                                                                  \\ \midrule
\makecell[c]{\textit{sta\_ask\_high\_priority\_ji}\\\textit{dur\_ask\_low\_priority\_ca}}          & \begin{tabular}[c]{@{}l@{}}What is your highest priority issue? Present your answer as one of the following multiple choice options. You must select an option.\\  A: position / B: company / C: salary / D: days\_off / E: workplace\end{tabular}                                                                                                                                                                               \\ \midrule
\makecell[c]{\textit{sta\_ask\_high\_priority\_ca}\\\textit{dur\_ask\_high\_priority\_ca}}          & \begin{tabular}[c]{@{}l@{}}What is your highest priority issue? Present your answer as one of the following multiple choice options. You must select an option.\\  A: food / B: water / C: firewood\end{tabular}                                                                                                                                                                                                                 \\ \midrule
\textit{dur\_strategy\_ca}                     & \begin{tabular}[c]{@{}l@{}}Which negotiation strategies are employed in the utterance? Present your answer as a comma-separated list of strategies, contained in \textless{}answer\textgreater{}\textless{}/answer\textgreater tags \\ with no additional text.\end{tabular}                                                                                                                                                      \\ \midrule
\textit{dur\_partner\_ask\_low\_priority\_ji}  & \begin{tabular}[c]{@{}l@{}}What is the recruiter's lowest priority issue? Present your answer as one of the following multiple choice options. You must select an option.\\   A: position / B: company / C: salary / D: days\_off / E: workplace\end{tabular}                                                                                                                                                                     \\ \midrule
\textit{dur\_partner\_ask\_low\_priority\_ca}  & \begin{tabular}[c]{@{}l@{}}What is your partner's lowest priority issue? Present your answer as one of the following multiple choice options. You must select an option.\\   A: food / B: water / C: firewood\end{tabular}                                                                                                                                                                                                         \\ \midrule
\textit{dur\_partner\_ask\_high\_priority\_ji} & \begin{tabular}[c]{@{}l@{}}What is the recruiter's highest priority issue? Present your answer as one of the following multiple choice options. You must select an option.\\  A: position / B: company / C: salary / D: days\_off / E: workplace\end{tabular}                                                                                                                                                                      \\ \midrule
\textit{dur\_partner\_ask\_high\_priority\_ca} & \begin{tabular}[c]{@{}l@{}}What is your partner's highest priority issue? Present your answer as one of the following multiple choice options. You must select an option.\\  A: food / B: water / C: firewood\end{tabular}                                                                                                                                                                                                       \\ \midrule
\makecell[c]{\textit{dur\_gen\_resp\_dnd}\\\textit{dur\_gen\_resp\_ca}}                   & Given the recent dialogue history inside \textless{}dialogue\textgreater tags, generate your next response in the negotiation concisely, following a similar style as previous utterances.                                                                                                                                                                                                                                       \\ \midrule
% \textit{dur\_gen\_resp\_ca}                    & Given the recent dialogue history inside \textless{}dialogue\textgreater tags, generate your next response in the negotiation concisely, following a similar style as previous utterances.                                                                                                                                                                                                                                       \\ \midrule
\textit{dur\_full\_proposal\_dnd}              & \begin{tabular}[c]{@{}l@{}}How many items does the speaker get for each issue in the proposal delimited by the \textless{}utterance\textgreater tags? Present your answer as a JSON within\textless{}answer\textgreater \textless{}/answer\textgreater tags \\ with keys as issues (books, hats, and balls) and values as the corresponding answers. If the answer is not clear for an issue, pick your best guess.\end{tabular} \\ \midrule
\textit{dur\_full\_proposal\_cra}              & \begin{tabular}[c]{@{}l@{}}How many items does the speaker get for each issue in the proposal delimited by the \textless{}utterance\textgreater tags? Present your answer as a JSON within\textless{}answer\textgreater \textless{}/answer\textgreater tags \\ with keys as issues (painting, lamp, and record) and values as the corresponding answers. If the answer is not clear for an issue, output NA.\end{tabular}      \\ \midrule
\makecell[c]{\textit{dur\_dial\_act\_ji}\\\textit{dur\_dial\_act\_cra}}                    & Which dialogue acts are employed in the utterance delimited by the\textless{}utterance\textgreater tags? Present your answer as a Python list of the relevant options. At least one option applies.                                                                                                                                                                                                                              \\ \midrule
\textit{dur\_dial\_act\_dnd}                   & Which dialogue act is employed in the utterance contained in\textless{}utterance\textgreater tags? Present your answer as a single word.                                                                                                                                                                                                                                                                                         \\ \midrule
\textit{end\_partner\_deal\_satisfaction\_ca}  & \begin{tabular}[c]{@{}l@{}}How satisfied do you think your partner is with the negotiation outcome? Present your answer as one of the following multiple choice options. You must select an option.\\  A: extremely\_dissatisfied / B: slightly\_dissatisfied / C: undecided / D: slightly\_satisfied / E: extremely\_satisfied\end{tabular}                                                                                     \\ \midrule
\textit{end\_partner\_deal\_likeness\_ca}      & \begin{tabular}[c]{@{}l@{}}How much do you think your partner likes you? Present your answer as one of the following multiple choice options. You must select an option.\\  A: extremely\_dissatisfied / B: slightly\_dissatisfied / C: undecided / D: slightly\_satisfied / E: extremely\_satisfied\end{tabular}                                                                                                                \\ \midrule
\makecell[c]{\textit{end\_deal\_total\_dnd}\\\textit{end\_deal\_total\_ca}}                 & How many points did you get at the end of the negotiation? Present your answer as a single number with no additional text.                                                                                                                                                                                                                                              \\ \midrule
% \textit{end\_deal\_total\_ca}                  & How many points did you get at the end of the negotiation? Present your answer as a single number with no additional text.                                                                                                                                                                                                                                                                                                       \\ \midrule
\textit{end\_deal\_specifics\_ji}              & \begin{tabular}[c]{@{}l@{}}In the final deal, what value was agreed on for each issue? Present your answer as a JSON within \textless{}answer\textgreater \textless{}/answer\textgreater tags \\ with keys as issues (Company, Position, Workplace, Salary, Days\_off)  and values as the corresponding answers. If you are unsure, pick your best guess.\end{tabular}                                                           \\ \midrule
\textit{end\_deal\_specifics\_dnd}             & \begin{tabular}[c]{@{}l@{}}In the final deal, how many items of each issue did you get? Present your answer as a JSON within \textless{}answer\textgreater \textless{}/answer\textgreater tags \\ with keys as issues (books, hats, and balls) and values as the corresponding answers. If you are unsure, pick your best guess.\end{tabular}                                                                                    \\ \midrule
\textit{end\_deal\_specifics\_ca}              & \begin{tabular}[c]{@{}l@{}}In the final deal, how many items of each issue did you get? Present your answer as a json within \textless{}answer\textgreater \textless{}/answer\textgreater tags \\ with keys as issues (food, water, and firewood) and values as the corresponding answers. If you are unsure, pick your best guess.\end{tabular}                                                                                \\ \midrule
\textit{end\_deal\_satisfaction\_ca}           & \begin{tabular}[c]{@{}l@{}}How satisfied are you with the negotiation outcome? Present your answer as one of the following multiple choice options. You must select an option.\\  A: extremely\_dissatisfied / B: slightly\_dissatisfied / C: undecided / D: slightly\_satisfied / E: extremely\_satisfied\end{tabular}                                                                                                          \\ \midrule
\textit{end\_deal\_likeness\_ca}               & \begin{tabular}[c]{@{}l@{}}How much do you like your partner? Present your answer as one of the following multiple choice options.You must select an option.\\  A: extremely\_dissatisfied / B: slightly\_dissatisfied / C: undecided / D: slightly\_satisfied / E: extremely\_satisfied\end{tabular}                                                                                                                             \\
\end{tabular}}
\caption{Questions used in the prompts for all tasks.}
\label{tab:Questions_prompt}
\end{table*}

\section{Task-specific Analysis}
\label{append:LLM_evaluation_tasks}

\begin{figure*}[ht]
    \centering
    \includegraphics[width=0.5\linewidth]{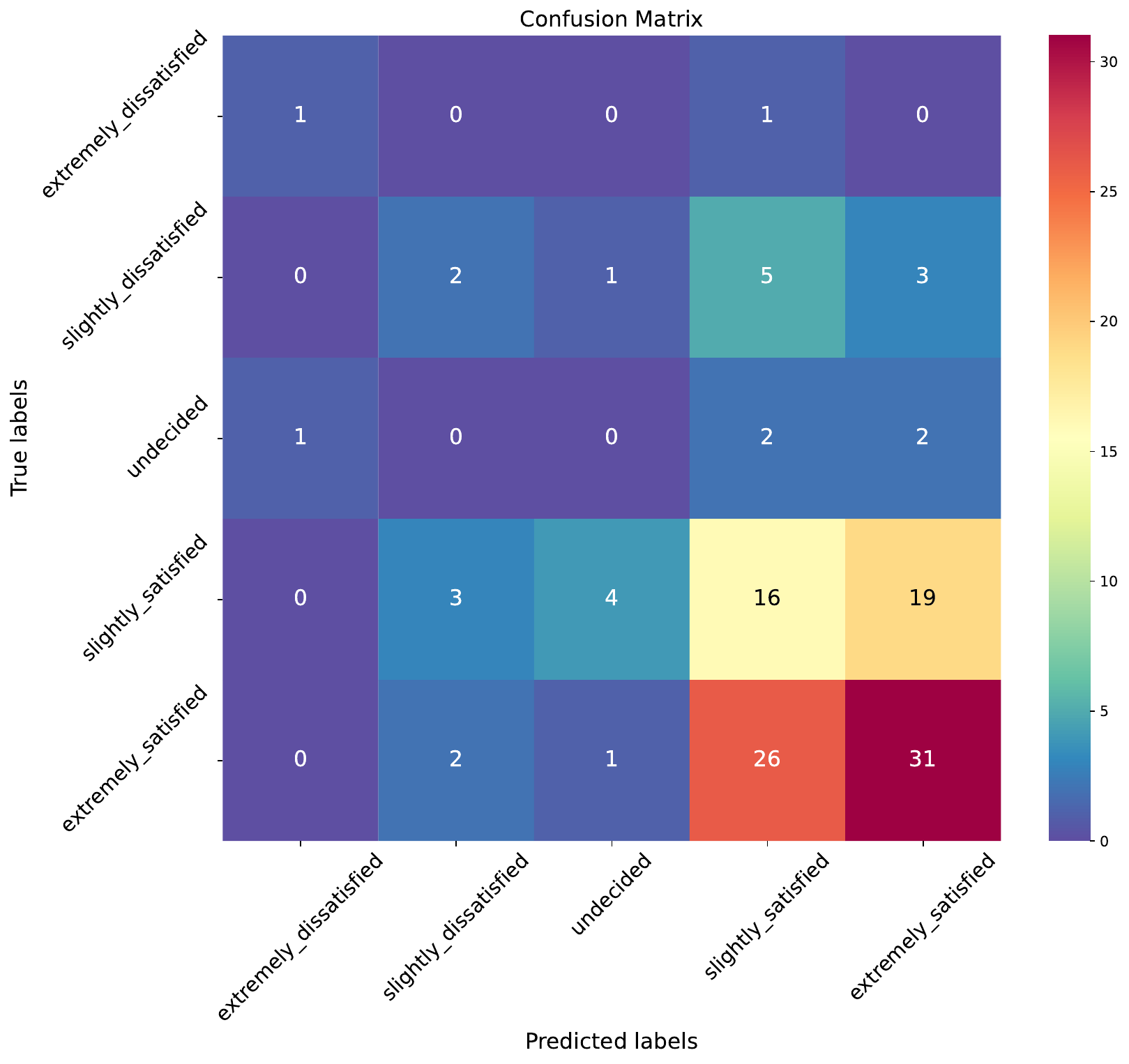}
    \caption{Confusion matrix of predictions of GPT-4 for the subjective task  (\textit{end\_deal\_satisfaction\_ca}).}
    \label{fig:confusion_append}
\end{figure*}

\begin{figure*}[ht]
    \centering
    \begin{subfigure}{0.49\textwidth}
        \includegraphics[width=\textwidth]{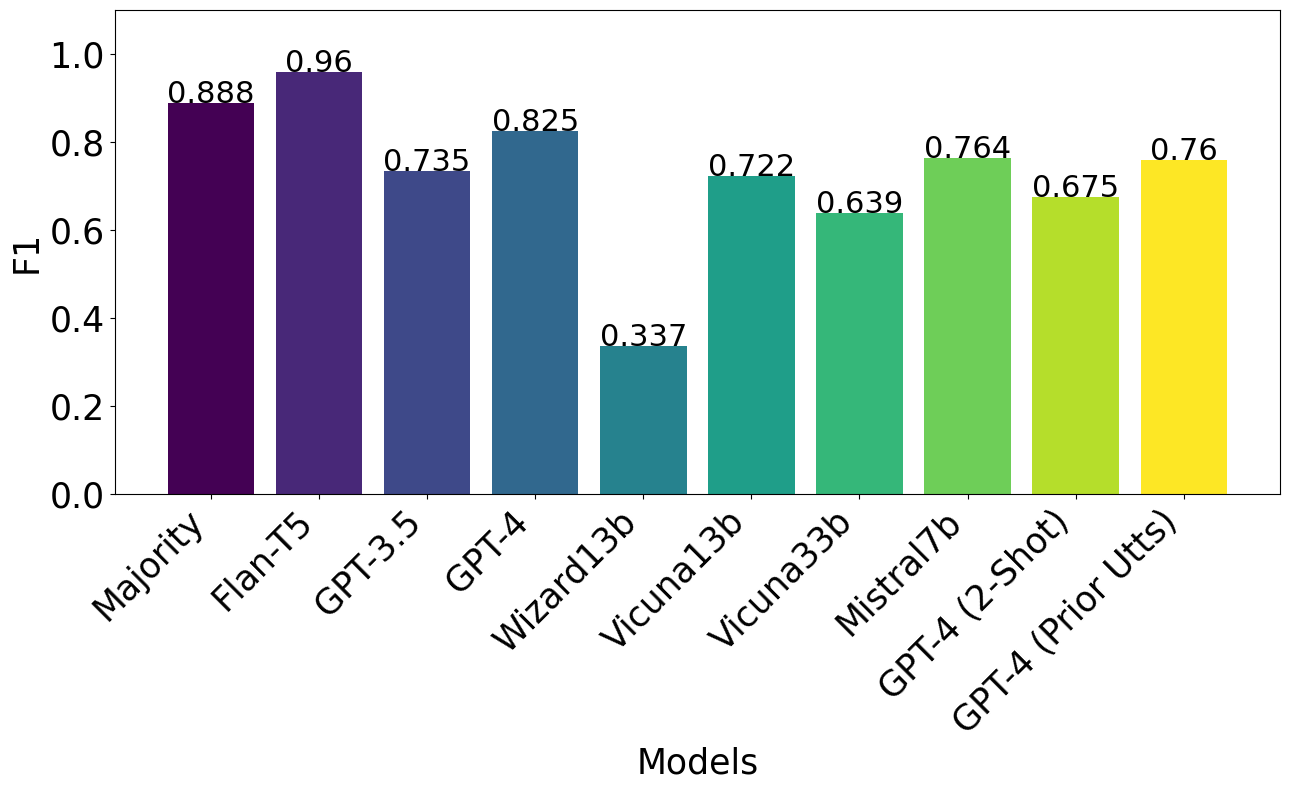} % Replace with the actual file name for your first figure
        \caption{Evaluation on \textit{dur\_dial\_act\_dnd} task}
    \end{subfigure}
    \hfill
    \begin{subfigure}{0.49\textwidth}
        \includegraphics[width=\textwidth]{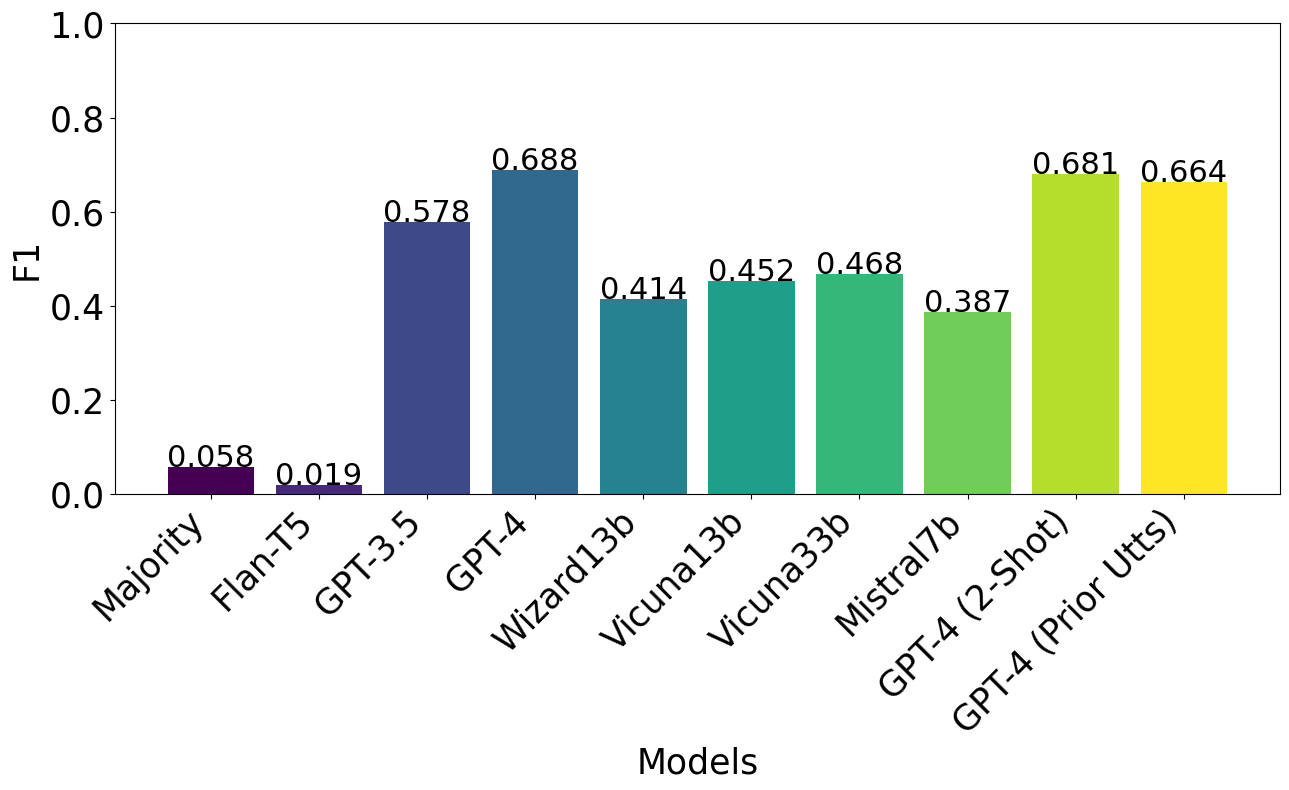} % Replace with the actual file name for your second figure
        \caption{Evaluation on \textit{dur\_dial\_act\_ji} task}
    \end{subfigure}
    \vskip\baselineskip
    \begin{subfigure}{0.49\textwidth}
        \includegraphics[width=\textwidth]{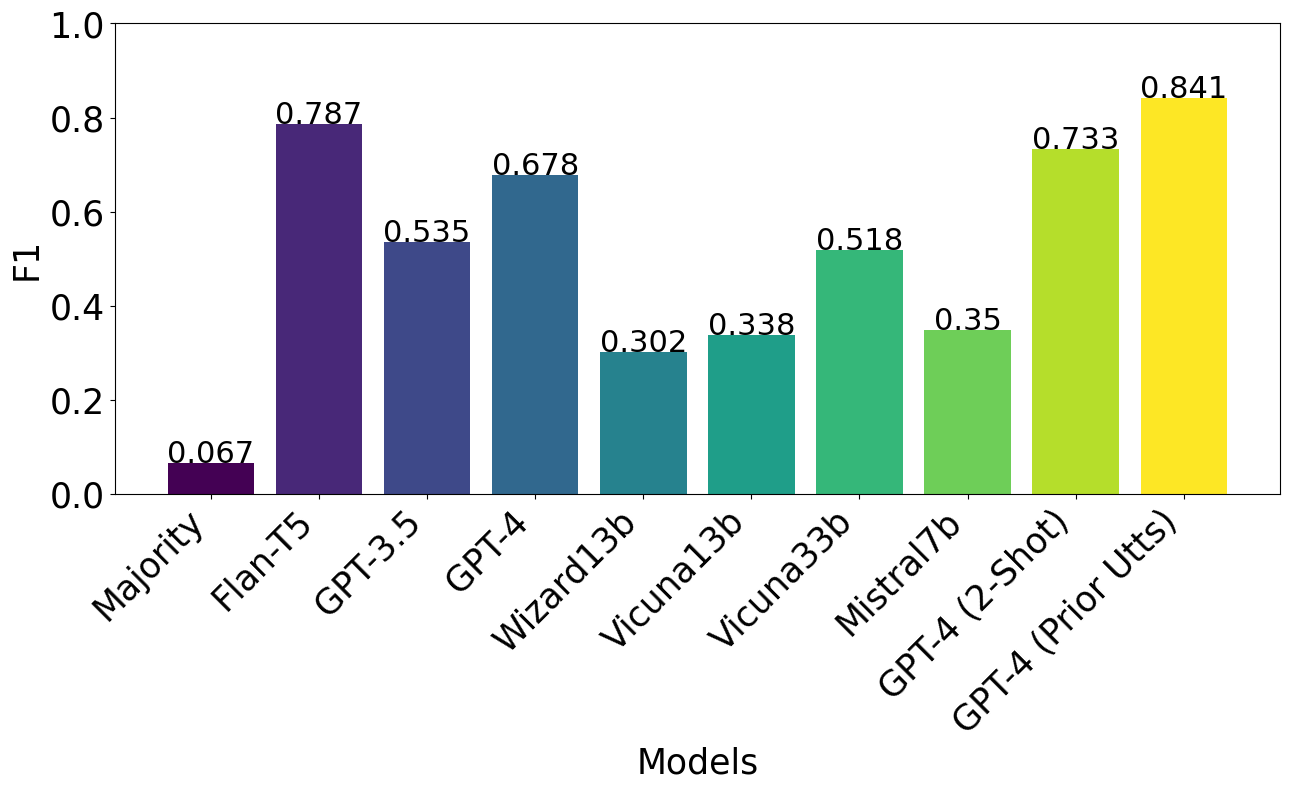} % Replace with the actual file name for your third figure
        \caption{Evaluation on \textit{dur\_dial\_act\_cra} task}
    \end{subfigure}
    \hfill
    \begin{subfigure}{0.49\textwidth}
        \includegraphics[width=\textwidth]{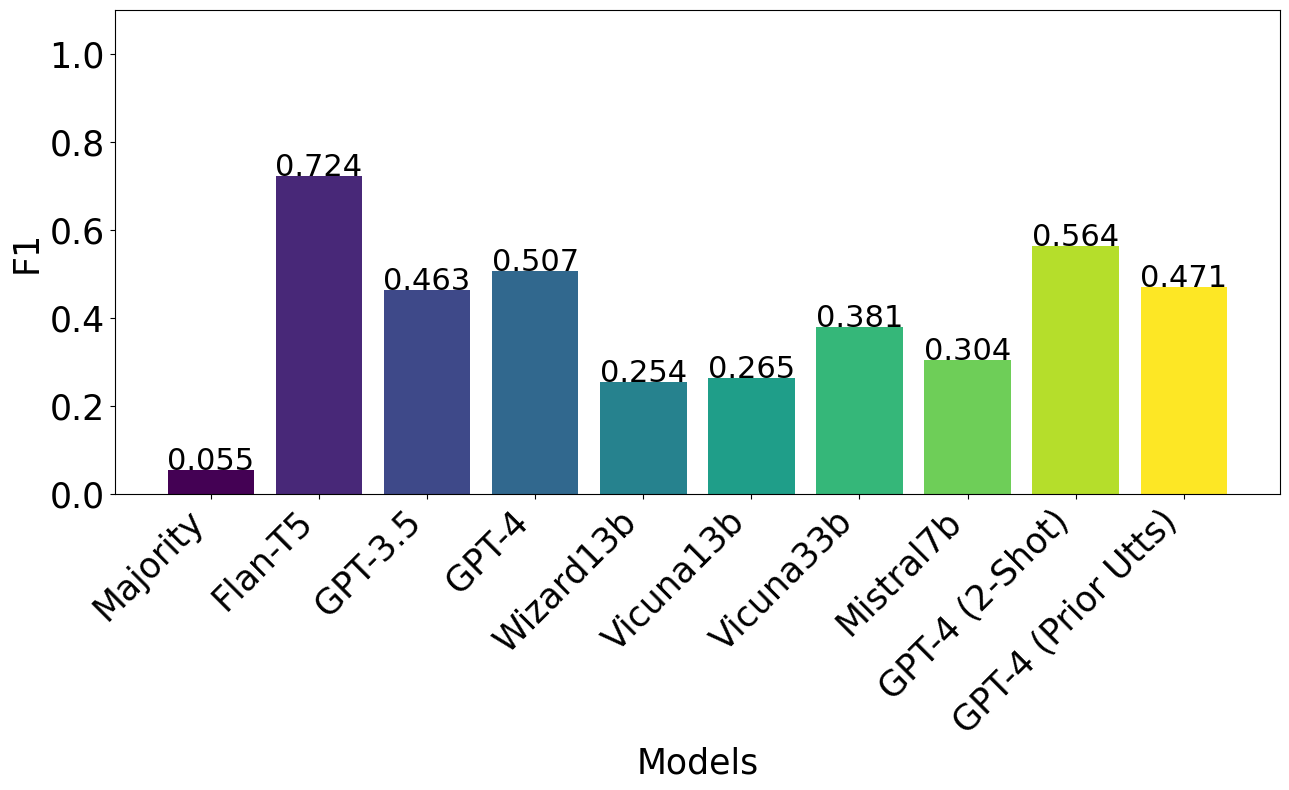} % Replace with the actual file name for your fourth figure
        \caption{Evaluation on \textit{dur\_strategy\_ca} task}
    \end{subfigure}
    \caption{Evaluation on four tasks demonstrating the impact of 1) Two-shot prompting and 2) Incorporating two prior utterances. We only consider GPT-4 for this analysis. F1 refers to the macro F1 score over all labels.}
    \label{fig:multi-shot-prompting}
\end{figure*}

\begin{figure*}[ht]
    \centering
    \begin{subfigure}{0.49\textwidth}
        \includegraphics[width=\textwidth]{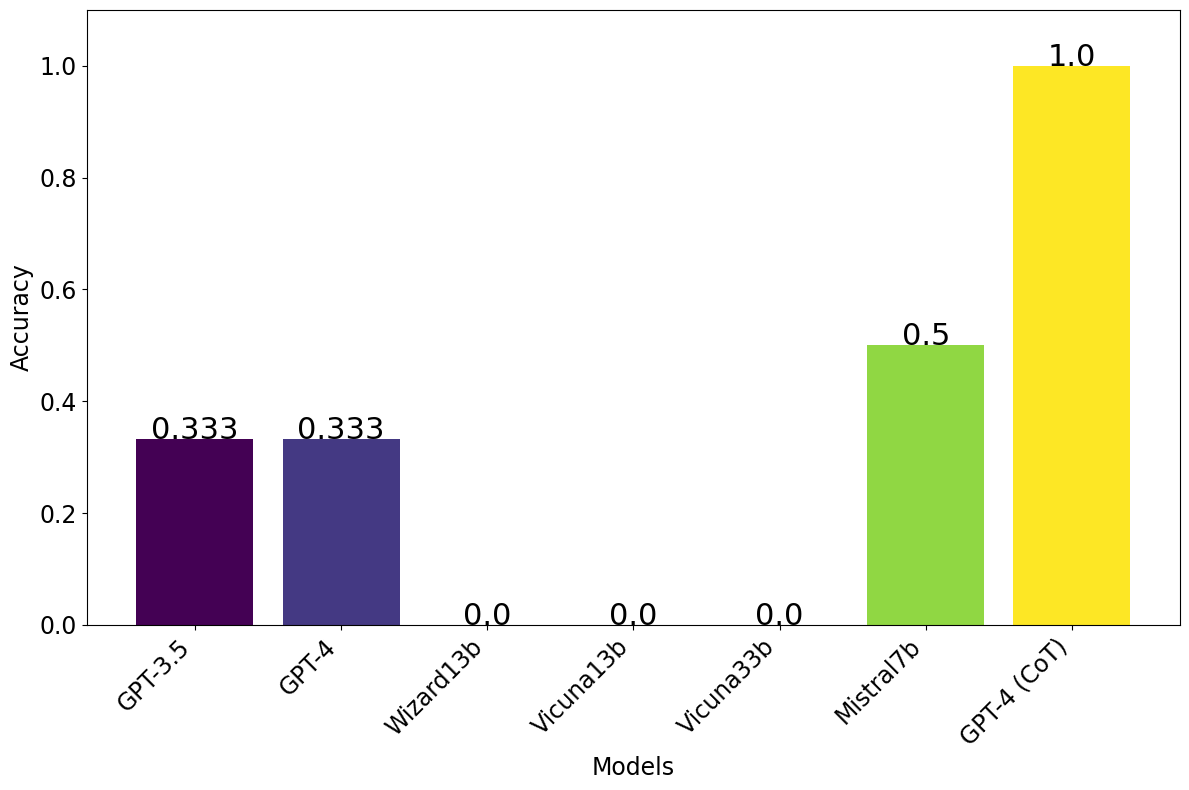} % Replace with the actual file name for your first figure
        \caption{Evaluation on \textit{sta\_max\_point\_ca} task}
        \label{fig:add_a}
    \end{subfigure}
    \hfill
    \begin{subfigure}{0.49\textwidth}
        \includegraphics[width=\textwidth]{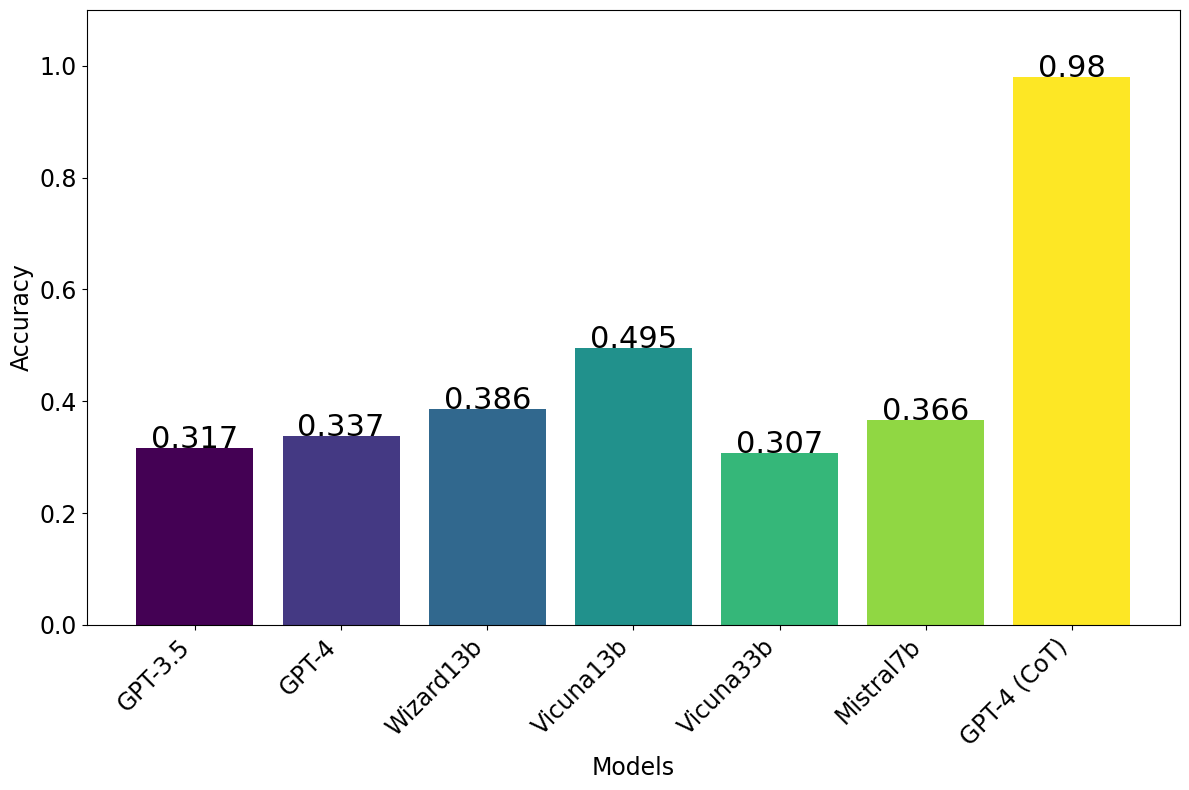} % Replace with the actual file name for your second figure
        \caption{Evaluation on  \textit{sta\_max\_point\_dnd} task}
        \label{fig:add_b}
    \end{subfigure}
    \vskip\baselineskip
    \begin{subfigure}{0.49\textwidth}
        \includegraphics[width=\textwidth]{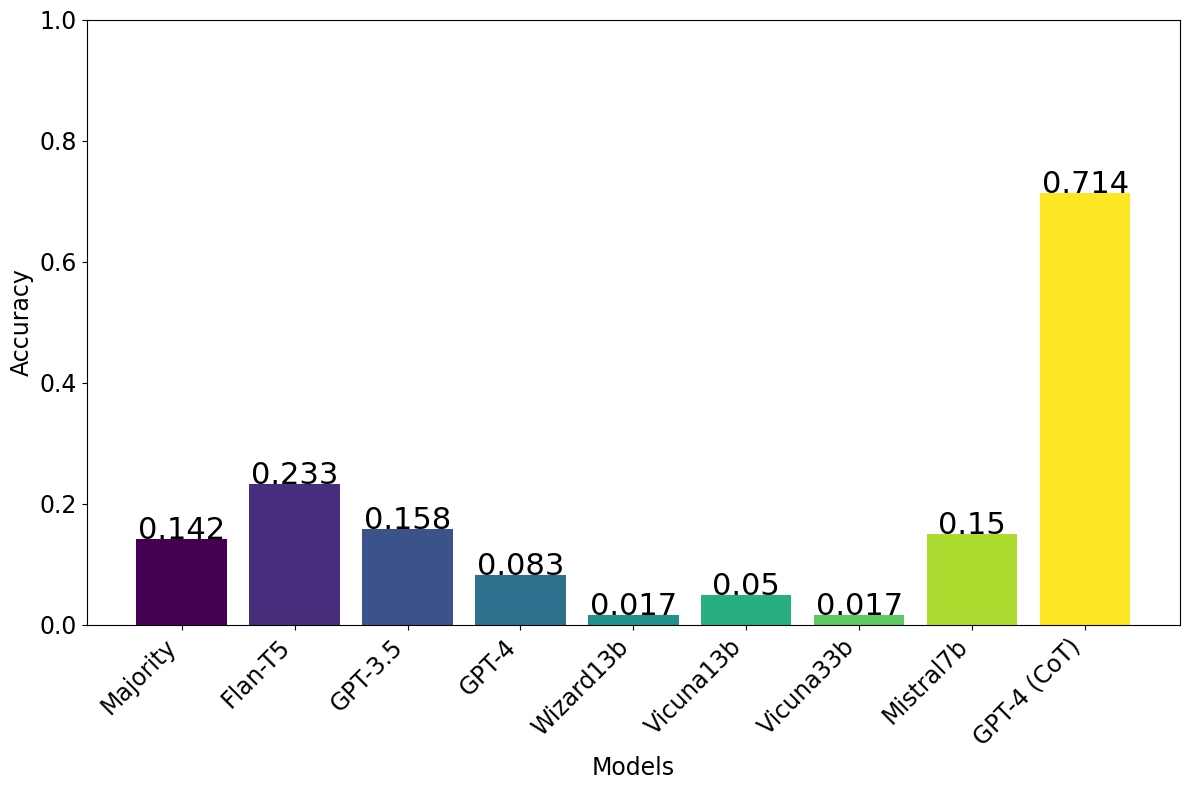} % Replace with the actual file name for your third figure
        \caption{Evaluation on \textit{end\_deal\_total\_ca} task}
        \label{fig:add_c}
    \end{subfigure}
    \hfill
    % \begin{subfigure}{0.49\textwidth}
    %     \includegraphics[width=\textwidth]{images/additional_end_deal_tot_dnd.png} % Replace with the actual file name for your fourth figure
    %     \caption{Evaluation on \textit{end\_deal\_total\_dnd} task}
    %     \label{fig:add_d}
    % \end{subfigure}
    \caption{Evaluation on tasks demonstrating the impact of CoT prompting. We only consider GPT-4 for this analysis.}
    \label{fig:CoT_prompting_append}
\end{figure*}

\begin{figure*}[ht]
    \centering
    \begin{subfigure}{0.49\textwidth}
        \includegraphics[width=\textwidth]{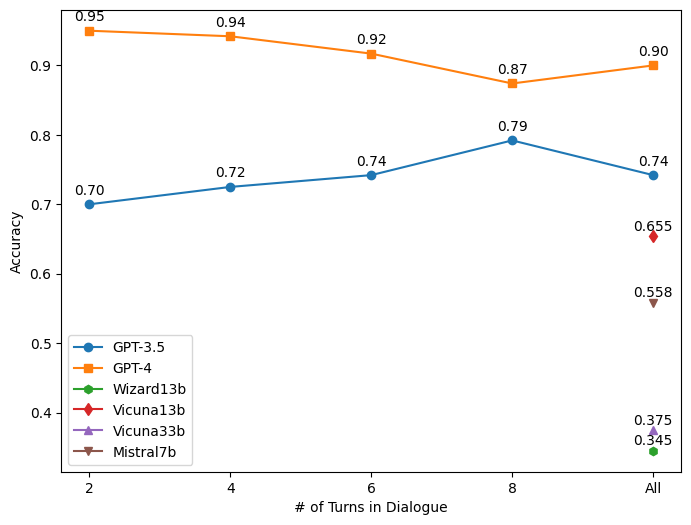} % Replace with the actual file name for your first figure
        \caption{Evaluation on \textit{dur\_ask\_high\_priority\_ca} task}
    \end{subfigure}
    \hfill
    \begin{subfigure}{0.49\textwidth}
        \includegraphics[width=\textwidth]{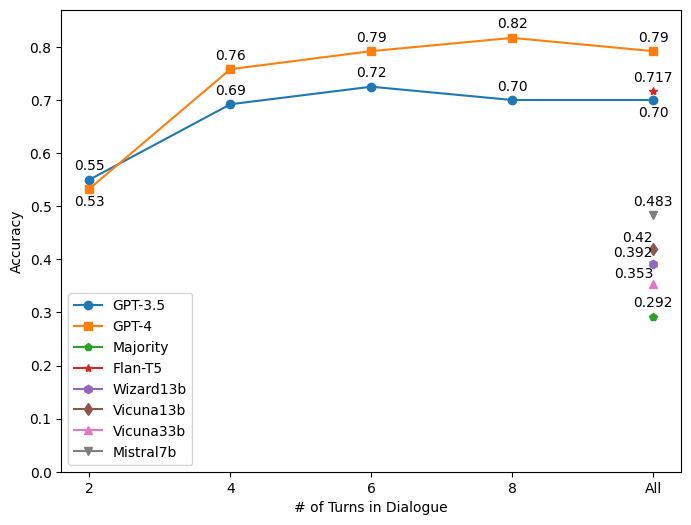} % Replace with the actual file name for your third figure
        \caption{Evaluation on \textit{dur\_partner\_ask\_high\_priority\_ca} task}
    \end{subfigure}
    \caption{Evaluation of two tasks not included in Figure ~\ref{fig:dynamic_priority}, illustrating the impact of increasing the number of utterances seen by GPT-3.5 and GPT-4 models.}
    \label{fig:num_of_dialogs}
\end{figure*}

Figure \ref{fig:confusion_append} displays the prediction results of GPT-4 on selected subjective tasks with a label-wise confusion matrix. Figure \ref{fig:multi-shot-prompting} presents the results for few-shot prompting and incorporating previous utterances for various \textit{Annotation} tasks. Figure \ref{fig:CoT_prompting_append} shows the results for the Chain-of-Thought (CoT) prompting technique on four multi-hop arithmetic reasoning \textit{Comprehension} tasks. We also present the effect of varying the number of utterances seen by the models for preference prediction tasks in Figure \ref{fig:num_of_dialogs}. 

Table \ref{tab:balance_test_tasks} shows the selected tasks in Table \ref{tab:KLD} presenting the evaluation results of LLMs on both well-balanced and imbalanced tasks.
\begin{table}[H]
\centering
\resizebox{\linewidth}{!}{%
\begin{tabular}{|l|l|}
\hline
\textbf{Task Type}        & \textbf{Tasks}                                                                                 \\ \hline
\textbf{Hard Priority Tasks}  & \textit{mid\_partner\_ask\_high/low\_priority\_ca}                                              \\ \hline
\textbf{Easy Priority Tasks}  & \textit{mid\_ask\_high/low\_priority\_ji}, \textit{sta\_ask\_high/low\_priority\_ji}            \\ \hline
\textbf{Dialog-Act/Strategy Tasks} & \textit{mid\_strategy\_ca, mid\_dial\_act\_cra, mid\_dial\_act\_dnd}                          \\ \hline
\end{tabular}}
\caption{Selected tasks categorized into Hard and Easy Priority Tasks, and Dialog-Act/Strategy Tasks.}
\label{tab:balance_test_tasks}
\end{table}

\section{Generation Outputs}
\label{sec:generation-samples-appendix}
We present sample generations by the compared models along with the ground-truth reference in Tables \ref{tab:sample_gen1} and \ref{tab:sample_gen2}.

\begin{table*}[ht]
\centering
\resizebox{\linewidth}{!}{%
\begin{tabular}{llcc}
Counts & Books: 1, Hats: 3, Balls: 2 & & \\
Agent Values & Each Book: 1 points, Each Hat: 1 points, Each Ball: 3 points & & \\ \midrule
Dialogue & \makecell[l]{THEM: i'm interested in the book and both balls in exchange for the hats .\\YOU: definitely not both balls . we could split them ?\\THEM: i think that's fair . i agree . 1 book , 1 ball for me . . . rest for you .} & & \\ \hline
& \makecell{Response} & Coherence & Strategy \\
Reference & YOU: deal ! & $5$ & $5$ \\
Flan-T5 & YOU: deal. & $5$ & $5$ \\ \hline
Mistral7b & \makecell[l]{YOU: I understand your offer, but I'm not willing to give up both balls. How about\\we split the balls and the hats evenly? That way, we can both get what we want.} & $2$ & $4$\\
Wizard13b & YOU: 1 book, 1 ball for me. & $1$ & $2$\\
Vicuna13b & YOU: 1 book, 1 ball for you, and the rest for me. & $2$ & $5$\\
Vicuna33b & YOU: Sounds good. I'll take the remaining book and both hats, and you can have the ball. & $2$ & $3$\\ \hline
GPT-3.5 & YOU: Sounds good. I'll take the 3 hats then. Deal. & $5$ & $5$ \\
GPT-4 & \makecell[l]{YOU: Great, so to confirm, I'll take 2 hats and 1 ball, and you get 1 book and 1 ball.\\Do we have a deal?} & $3$ & $4$\\
\end{tabular}}
\caption{Sample responses generated by the models for an instance from the DND dataset. The \textit{Coherence} and \textit{Strategy} scores are based on human annotations on a scale of $1$-$5$.}
\label{tab:sample_gen1}
\end{table*}

\begin{table*}[ht]
\centering
\resizebox{\linewidth}{!}{%
\begin{tabular}{llcc}
Counts & Food Packages: 3, Water Packages: 3, Firewood Packages: 3 & & \\
Agent Values & Each Food Package: 3 points, Each Water Package: 4 points, Each Firewood Package: 5 points & & \\ \midrule
Dialogue & \makecell[l]{YOU: hello\\THEM: Hello, there. Should we discuss how to divide up the items for the camping trip?\\YOU: yeah. i need 2 firewood, 1 water and 1 food\\THEM: I need to have 2 or 3 pieces of firewood because I did not pack any way to\\heat up my food without firewood.} & & \\ \hline
& \makecell{Response} & Coherence & Strategy \\
Reference & YOU: I also need 2 pieces of firewood & $4$ & $3$ \\
Flan-T5 & YOU: i need 2 firewood, 2 water and 1 food & $3$ & $3$ \\ \hline
Mistral7b & \makecell[l]{YOU: Based on the recent dialogue history, it seems like we both need firewood for\\our camping trip. How about we agree to split the 3 firewood packages equally,\\and I'll take the 1 water and 1 food package?} & $5$ & $4$\\
Wizard13b & \makecell[l]{YOU: I understand, I can provide you with 2 firewood, 1 water, and 1 food. How\\about we split the remaining items equally?} & $5$ & $3$\\
Vicuna13b & \makecell[l]{YOU: I understand your need for firewood. How about we split the firewood packages\\evenly, with each of us getting 1 and a half packages? That way, you'll have enough firewood\\for your needs, and I'll still have some for my camping trip.} & $5$ & $3$\\
Vicuna33b & \makecell[l]{YOU: I understand your need for firewood. Let's propose a deal: I'll take 1 firewood,\\1 water, and 2 food packages. In return, you can have the remaining 2 firewood\\packages. This way, we both get what we need.} & $5$ & $2$\\ \hline
GPT-3.5 & YOU: I can give you 2 firewood, 1 water, and 1 food package. & $4$ & $5$ \\
GPT-4 & \makecell[l]{YOU: How about if I take 2 firewood and you take 1, and in exchange, I can let you\\have 2 water packages?} & $4$ & $4$\\
\end{tabular}}
\caption{Sample responses generated by the models for an instance from the CaSiNo dataset. The \textit{Coherence} and \textit{Strategy} scores are based on human annotations on a scale of $1$-$5$.}
\label{tab:sample_gen2}
\end{table*}

\end{document}